\documentclass{article}

\usepackage[nonatbib,final]{neurips_2021}

\PassOptionsToPackage{usenames,dvipsnames}{xcolor}
\usepackage{xcolor}

\usepackage[utf8]{inputenc} 
\usepackage[T1]{fontenc}    
\usepackage{hyperref}       
\usepackage{url}            
\usepackage{booktabs}       
\usepackage{amsfonts}       
\usepackage{nicefrac}       
\usepackage{microtype}      
\usepackage{graphicx}
\usepackage[square, numbers]{natbib}
\usepackage{amsmath}
\usepackage{amsthm}
\usepackage{mathtools}
\usepackage{amssymb}
\usepackage{todonotes}
\usepackage[makeroom]{cancel}
\usepackage{caption}
\usepackage{subcaption}
\usepackage{tikz}
\usepackage{cancel}

\usepackage{algorithm}
\usepackage{algorithmic}

\usepackage{mwe} 

\usepackage[capitalise,nameinlink]{cleveref}

\newcommand{\basemodel}{f_{\mathbf{w}_*}}
\newcommand{\oldreg}{\mathcal{R}}
\newcommand{\newreg}{\mathcal{G}}
\newcommand{\modelout}{f_{\text{\vparam}}}
\newcommand{\memory}{\mathcal{M}}
\newcommand{\kprior}{\mathcal{K}}

\newcommand{\inspace}{\mathcal{X}}
\newcommand{\data}{\mathcal{D}}

\newcommand{\breg}{\mathcal{B}}

\newcommand{\barloss}{\bar{\ell}}

\newcommand{\loss}{\ell}

\newcommand{\vparam}{\vw}
\newcommand{\param}{w}

\newcommand{\comment}[1]{}

\newcommand{\dkls}[3]{\mathbb{D}_{\text{KL}}^{#1}[#2 \, \|\, #3]}

\newcommand\cut[1]{}

\newcommand{\squishlist}{
   \begin{list}{$\bullet$}
    { \setlength{\itemsep}{0pt}      \setlength{\parsep}{3pt}
      \setlength{\topsep}{3pt}       \setlength{\partopsep}{0pt}
      \setlength{\leftmargin}{1.5em} \setlength{\labelwidth}{1em}
      \setlength{\labelsep}{0.5em} } }

\newcommand{\squishlisttwo}{
   \begin{list}{$\bullet$}
    { \setlength{\itemsep}{0pt}    \setlength{\parsep}{0pt}
      \setlength{\topsep}{0pt}     \setlength{\partopsep}{0pt}
      \setlength{\leftmargin}{2em} \setlength{\labelwidth}{1.5em}
      \setlength{\labelsep}{0.5em} } }

\newcommand{\squishend}{
    \end{list}  }

{}
\newtheorem{thm}{Theorem}{}
{}
{}
{}

{}

{}

\newcommand{\half}{\mbox{$\frac{1}{2}$}}

\newcommand{\real}{\mbox{$\mathbb{R}$}}

\newcommand{\rnd}[1]{\left(#1\right)}
\newcommand{\sqr}[1]{\left[#1\right]}

\newcommand{\myang}[1]{\langle#1\rangle}
\newcommand{\myexpect}{\mathbb{E}}

\newcommand{\gauss}{\mbox{${\cal N}$}}

\newcommand{\myvec}[1]{\mbox{$\mathbf{#1}$}}
\newcommand{\myvecsym}[1]{\mbox{$\boldsymbol{#1}$}}

\newcommand{\valpha}{\mbox{$\myvecsym{\alpha}$}}
\newcommand{\vbeta}{\mbox{$\myvecsym{\beta}$}}

\newcommand{\veta}{\mbox{$\myvecsym{\eta}$}}

\newcommand{\vmu}{\mbox{$\myvecsym{\mu}$}}

\newcommand{\vlambda}{\mbox{$\myvecsym{\lambda}$}}
\newcommand{\vLambda}{\mbox{$\myvecsym{\Lambda}$}}

\newcommand{\vphi}{\mbox{$\myvecsym{\phi}$}}
\newcommand{\vPhi}{\mbox{$\myvecsym{\Phi}$}}

\newcommand{\vtheta}{\mbox{$\myvecsym{\theta}$}}

\newcommand{\vSigma}{\mbox{$\myvecsym{\Sigma}$}}

\newcommand{\va}{\mbox{$\myvec{a}$}}

\newcommand{\vd}{\mbox{$\myvec{d}$}}
\newcommand{\ve}{\mbox{$\myvec{e}$}}

\newcommand{\vf}{\mbox{$\myvec{f}$}}

\newcommand{\vk}{\mbox{$\myvec{k}$}}
\newcommand{\vm}{\mbox{$\myvec{m}$}}

\newcommand{\vu}{\mbox{$\myvec{u}$}}
\newcommand{\vv}{\mbox{$\myvec{v}$}}
\newcommand{\vw}{\mbox{$\myvec{w}$}}

\newcommand{\vx}{\mbox{$\myvec{x}$}}

\newcommand{\vy}{\mbox{$\myvec{y}$}}

\newcommand{\vA}{\mbox{$\myvec{A}$}}
\newcommand{\vB}{\mbox{$\myvec{B}$}}

\newcommand{\vD}{\mbox{$\myvec{D}$}}

\newcommand{\vG}{\mbox{$\myvec{G}$}}

\newcommand{\vI}{\mbox{$\myvec{I}$}}

\newcommand{\vK}{\mbox{$\myvec{K}$}}

\newcommand{\vS}{\mbox{$\myvec{S}$}}

\newcommand{\vU}{\mbox{$\myvec{U}$}}
\newcommand{\vV}{\mbox{$\myvec{V}$}}

\newcommand{\be}{\begin{equation}}
\newcommand{\ee}{\end{equation}}
\newcommand{\bea}{\begin{eqnarray}}
\newcommand{\eea}{\end{eqnarray}}
\newcommand{\beaa}{\begin{eqnarray*}}
\newcommand{\eeaa}{\end{eqnarray*}}

\crefname{section}{Sec.}{Sections}
\crefname{appendix}{App.}{Appendices}
\crefname{algorithm}{Algo.}{Algorithms}
\crefname{equation}{Eq.}{Equations}
\crefname{figure}{Fig.}{Figures}
\creflabelformat{equation}{#2\textup{#1}#3}
\usepackage{amsmath}

\DeclareMathOperator*{\argmin}{arg\,min}

\title{Knowledge-Adaptation Priors}

\author{Mohammad Emtiyaz Khan$^*$\\
RIKEN Center for AI Project\\ 
Tokyo, Japan\\
\tt{emtiyaz.khan@riken.jp}
\And
Siddharth Swaroop$^*$\\
University of Cambridge\\ 
Cambridge, UK \\
\tt{ss2163@cam.ac.uk}
}

\begin{document}

\maketitle
\let\thefootnote\relax\footnotetext{* Authors contributed equally.}

\begin{abstract}
   Humans and animals have a natural ability to quickly adapt to their surroundings, but machine-learning models, when subjected to changes, often require a complete retraining from scratch.~We present Knowledge-adaptation priors (K-priors) to reduce the cost of retraining by enabling quick and accurate adaptation for a wide-variety of tasks and models.~This is made possible by a combination of weight and function-space priors to reconstruct the gradients of the past, which recovers and generalizes many existing, but seemingly-unrelated, adaptation strategies.
   Training with simple first-order gradient methods can often recover the exact retrained model to an arbitrary accuracy by choosing a sufficiently large memory of the past data.
   Empirical results show that adaptation with K-priors achieves performance similar to full retraining, but only requires training on a handful of past examples.
\end{abstract}

\section{Introduction}
Machine-Learning (ML) at production often requires constant model updating which can have huge financial and environmental costs \citep{diethe2019continual, Paleyes_Urma_Lawrence_2021}. The production pipeline is continuously evolving, where new data are regularly pooled and labeled and old data become irrelevant. Regular tuning of hyperparameters is required to handle drifts~\citep{diethe2019continual}, and sometimes even the model class/architecture may need to change.
Due to this, the model is frequently retrained, retested, and redeployed, which can be extremely costly, especially when the data and model sizes are large. 
The cost can be reduced if, instead of repeated retraining, the system can quickly adapt to incremental changes. Humans and animals can naturally use their prior knowledge to handle a wide-variety of changes in their surroundings, but such quick, wide, and accurate adaptation has been difficult to achieve in ML.

In theory, this should be possible within a Bayesian framework where the posterior is used as the prior for the future, but exact Bayes is computationally challenging and the design of generic Bayesian priors has its own challenges \citep{simpson2017, nalisnick2021predictive}.
In ML, simpler mechanisms are more popular, for example, in Support Vector Machines (SVMs) for adding/removing data~\citep{cauwenberghs2001incremental, Tsai_Lin_Lin_2014, vapnik2015learning}, and in deep learning for model compression~\citep{hinton2015distilling}.
Weight-priors are used in online learning~\citep{cesa2006prediction}, and more recently for continual learning~\citep{kirkpatrick2017overcoming, nguyen2017variational, ritter2018online, lee2017overcoming, zenke2017continual}, but they are not suited for many other tasks, such as model compression. In some settings, they also perform worse, for example, in continual learning when compared to memory-based strategies~\citep{li2017learning, rebuffi2017icarl, pan2020continual, buzzega2020dark}. 
All these previous works apply to narrow, specific settings, and designing generic adaptation-mechanisms remains an open challenge. 

\begin{figure}[!t]
   \includegraphics[width=\textwidth]{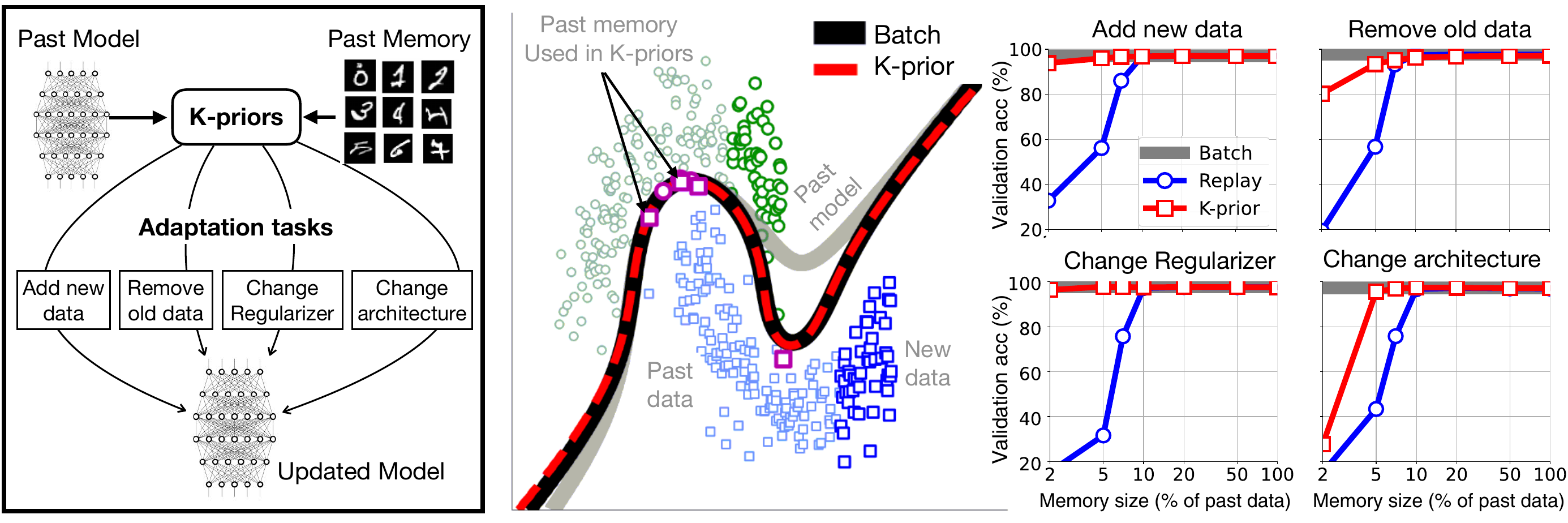}
   \caption{\textbf{Left:} K-priors can handle a wide-variety of adaptation tasks by using the past model and a small memory of the past data. \textbf{Middle:} For adaptation, the model needs tweaking at only a handful of past examples (shown with dark purple markers). The adapted model (dashed red line) is very close to the retraining on the full batch (solid black line). \textbf{Right:} Results on binary classification on `USPS' digits with neural networks show that K-priors (red square)
   obtain solutions close to the batch training (gray line) but by using only a fraction (2-5\%) of the past data. It also performs better than `Replay' (blue circles) where the same memory is used for replay. See \cref{sec:experiments} for more details.}
   \label{fig:fig1}
\end{figure}

We present Knowledge-adaptation priors (K-priors) for the design of generic adaptation-mechanisms. The general principle of adaptation is to combine the weight and function-space divergences to faithfully reconstruct the gradient of the past.
K-priors can handle a wide variety of adaptation tasks (\cref{fig:fig1}, left) and work for a range of models, such as generalized linear models, deep networks, and their Bayesian
extensions.
The principle unifies and generalizes many seemingly-unrelated existing works, for example,  weight-priors~\cite{cesa2006prediction}, knowledge distillation~\cite{hinton2015distilling}, SVMs~\citep{cauwenberghs2001incremental, Liang_Li_2009, Tsai_Lin_Lin_2014}, online Gaussian Processes~\citep{csato2002sparse}, and continual learning~\cite{lopez2017gradient, pan2020continual, buzzega2020dark}.   
It leads to natural adaptation-mechanisms where models' predictions need to be readjusted only at a handful of past experiences (\cref{fig:fig1}, middle). This is quick and easy to train with first-order optimization methods and, by choosing a sufficiently large memory, we obtain results similar to retraining-from-scratch (\cref{fig:fig1}, right).
Code is available at \texttt{\href{https://github.com/team-approx-bayes/kpriors}{https://github.com/team-approx-bayes/kpriors}}.

\section{Adaptation in Machine Learning}
\label{sec:background}

\subsection{Knowledge-adaptation tasks}
Our goal is to quickly and accurately adapt an already trained model to incremental changes in its training framework. Throughout we refer to the trained model as the \emph{base model}. We denote its outputs by $\basemodel(\vx)$ for inputs $\vx\in\real^D$, and assume that its parameters $\vparam_*\in\mathcal{W} \subset\real^P$ are obtained by solving the following problem on the data $\data$,
\begin{equation}
   \vparam_* = \argmin_{\text{\vparam}\in\mathcal{W}} \,\, \barloss(\vparam), \quad \text{ where } \barloss(\vparam) = \sum_{i\in\data} \loss_i(\vparam) + \oldreg(\vparam).
   \label{eq:ERMold}
\end{equation}
Here, $\loss_i(\vparam)$ denotes the loss function on the $i$'th data example, and $\oldreg(\vparam)$ is a regularizer. 

A good adaptation method should be able to handle many types of changes. The simplest and most common change is to add/remove data examples, as shown below at the left where the data example $j\notin \data$ is added to get $\vparam_+$, and at the right where an example $k\in\data$ is removed to get $\vparam_-$,
\begin{align}
   \vparam_{+} &= \argmin_{\text{\vparam}\in\mathcal{W}} \,\, \sum_{i\in\data\cup j} \loss_i(\vparam) + \oldreg(\vparam), 
   &&\quad \vparam_{-} = \argmin_{\text{\vparam}\in\mathcal{W}} \,\, \sum_{i\in\data \backslash k} \loss_i(\vparam) + \oldreg(\vparam).
   \label{eq:addremove}
\end{align}
We refer to these problems as `Add/Remove Data' tasks.

Other changes are the `Change Regularizer' (shown on the left) task where the regularizer is replaced by a new one $\newreg(\vparam)$, and the `Change Model Class/Architecture' task (shown on the right) where the model class/architecture is changed leading to a change of the parameter space from $\mathcal{W}$ to $\Theta$,
\begin{align}
   \vparam_{\newreg} &= \argmin_{\text{\vparam}\in\mathcal{W}} \,\, \sum_{i\in\data} \,\,\,\, \loss_i(\vparam) + \newreg(\vparam), 
   &&\quad \vtheta_* = \argmin_{\text{\vtheta}\in \Theta} \,\, \sum_{i\in\data} \,\,\,\, \tilde{\loss}_i(\vtheta) + \tilde{\oldreg}(\vtheta) .
   \label{eq:reg_compress}
\end{align}
Here, we assume the loss $\tilde{\loss}_i(\vtheta)$ has the same form as $\loss_i(\vparam)$ but using the new model $\tilde{f}_{\text{\vtheta}}(\vx)$ for prediction, with $\vtheta \in \Theta$ as the new parameter. $\vtheta$ can be of a different dimension, and the new regularizer can be chosen accordingly.  
The change in the model class could be a simple change in features, for example it is common in linear models to add/remove features, or the change could be similar to model compression or knowledge distillation \citep{hinton2015distilling}, which may not have a regularizer.

Another type of change is to add \emph{privileged} information, originally proposed by \citet{vapnik2015learning}. The goal is to include different types of data to improve the performance of the model. This is combined with knowledge distillation by \citet{lopez2015unifying} and has recently been applied to domain adaptation in deep learning~\citep{ao2017fast, ruder2017knowledge, sarafianos2017adaptive}. 

There could be several other kinds of changes in the training framework, for example, those involving a change in the loss function, or even a combination of the changes discussed above. Our goal is to develop a method that can handle such wide-variety of changes, or `adaptation tasks' as we will call them throughout. 
Such adaptation can be useful to reduce the cost of model updating, for example in a continuously evolving ML pipeline.
Consider $k$-fold cross-validation, where the model is retrained from scratch for every data-fold and hyperparameter setting.
Such retraining can be made cheaper and faster by reusing the model trained in the previous folds and adapting them for new folds and hyperparameters.
Model reuse can also be useful during active-learning for dataset curation, where a decision to include a new example can be made by using a quick Add Data adaptation.
In summary, model adaptation can reduce the cost by avoiding the need to constantly retrain.

\subsection{Challenges of knowledge adaptation}
Knowledge adaptation in ML has proven to be challenging.
Currently there are no methods that can handle many types of adaptation tasks.
Most existing works apply narrowly to specific models and mainly focuses on adaptation to Add/Remove Data only. 
This includes many early proposals for SVMs \citep{cauwenberghs2001incremental, tveit2003incremental, Karasuyama_Takeuchi_2010, duan2007decremental, romero2007incremental, Liang_Li_2009, laskov2006incremental, Tsai_Lin_Lin_2014}, recent ones for machine-unlearning~\citep{Cao_Yang_2015, golatkar2020eternal, Nguyen_Low_Jaillet_2020,
guo2019certified, schelter2019amnesia, brophy21}, and methods that use weight and functional priors, for example, for online learning \cite{cesa2006prediction}, continual deep learning~\cite{kirkpatrick2017overcoming, nguyen2017variational, ritter2018online, lee2017overcoming, zenke2017continual, li2017learning, rebuffi2017icarl, benjamin2018measuring,
titsias2020functional, pan2020continual, buzzega2020dark}, and Gaussian-Process (GP) models \citep{csato2002sparse, sarkka2013spatiotemporal, Solin_Hensman_Turner_2018}.
The methodologies of these methods are entirely different from each other, and they do not appear to have any common principles behind their adaptation mechanisms. Our goal is to fill this gap and propose a single method that can handle a wide-variety of tasks for a range of models.

We also note that there is no prior work on the Change Regularizer task. Adaptation has been used to boost hyperparameter tuning~\citep{wen2017improving} but only Add/Remove Data adaptation is used. Warm-starts have been employed as well~\cite{DeCoste_Wagstaff_2000}, but it is often not sufficient and can even hurt performance~\citep{NEURIPS2020_288cd256}.

\subsection{Problem setting and notation}
Throughout, we will use a supervised problem where the loss is specified by an exponential-family,
\begin{align}
   \loss(y, h(f)) = - \log p(y|f) = - \myang{y, f} + A(f),  
   \label{eq:loss_func}
\end{align}
where $y\in\mathcal{Y}$ denotes the scalar observation output, $f\in\mathcal{F}$ is the canonical natural parameter, $A(f)$ is the log-partition function, and $h(f) = \myexpect(y) = \nabla A(f)$ is the expectation parameter.
A typical example is the cross-entropy loss for binary outcomes $y\in\{0,1\}$ where $A(f) = \log (1+ e^f)$ and $h(f) = \sigma(f)$ is the Sigmoid function.
It is straightforward to extend our method to a vector observation and model outputs.
An extension to other types of learning frameworks is discussed in the next section (see the discussion around \cref{eq:genericKpriors}). 

Throughout the paper, we will use a shorthand for the model outputs, where we denote $\modelout^i = \modelout(\vx_i)$. We will repeatedly make use of the following expression for the derivative of the loss,
\begin{align}
   \nabla \loss(y_i, h(\modelout^i)) = \nabla \modelout^i \,\, [ h(\modelout^i) - y_i].
   \label{eq:grad_loss}
\end{align}

\section{Knowledge-Adaptation Priors (K-priors)}
\label{sec:kpriors}

We present Knowledge-adaptation priors (K-priors) to quickly and accurately adapt a model's knowledge to a wide variety of changes in its training framework.
K-priors, denoted below by $\kprior(\vparam;\vparam_*, \memory)$, refer to a class of priors that use both weight and function-space regularizers,
\begin{align}
   \kprior(\vparam; \vparam_*, \memory) &=  \mathbb{D}_f\rnd{\vf(\vparam)\| \vf(\vparam_*)} + \tau \mathbb{D}_{\param} \rnd{\vparam \| \vparam_*},
   \label{eq:kpost_def}
\end{align}
where $\vf(\vparam)$ is a vector of $\modelout(\vu_i)$, defined at inputs in $\memory = (\vu_1, \vu_2, \ldots, \vu_M)$. The divergence $\mathbb{D}_f(\cdot\|\cdot)$ measures the discrepancies in the function space $\mathcal{F}$, while $\mathbb{D}_\param(\cdot\|\cdot)$ measures the same in the weight space $\mathcal{W}$. Throughout, we will use Bregman divergences 
$\breg_\psi(p_1, p_2) = \psi(p_1) - \psi(p_2) - \nabla \psi(p_2)^\top (p_1 - p_2)$, 
specified using a strictly-convex Bregman function $\psi(\cdot)$.

K-priors are defined using the base model $\vparam_*$, the memory set $\memory$, and a trade-off parameter $\tau>0$. We keep $\tau=1$ unless otherwise specified. It might also use other parameters required to define the divergence functions. 
We will sometimes omit the dependency on parameters and refer to $\kprior(\vparam)$.

Our general principle of adaptation is to use $\kprior(\vparam)$ to faithfully reconstruct the gradients of the past training objective. This is possible due to the combination of weight and function-space divergences. Below, we illustrate this point for supervised learning for Generalized Linear Models (GLMs).

\subsection{K-priors for GLMs}
GLMs include models such as logistic and Poisson regression, and have a linear model $\modelout^i = \vphi_i^\top \vparam$, with feature vectors $\vphi_i = \vphi(\vx_i)$.
The base model is obtained as follows,
\begin{equation}
   \vparam_* = \argmin_{\text{\vparam}\in\mathcal{W}} \,\, \sum_{i\in\data} \loss(y_i, h(\modelout^i)) + \oldreg(\vparam).
   \label{eq:GLMobjective}
\end{equation}
In what follows, for simplicity, we use an $L_2$ regularizer $\oldreg(\vparam) = \half \delta \|\vparam\|^2$, with $\delta>0$.

We will now discuss a K-prior that \emph{exactly} recovers the gradients of this objective.
For this, we choose $\mathbb{D}_\param(\cdot\|\cdot)$ to be the Bregman divergence with $\oldreg(\vparam)$ as the Bregman function, 
\[\mathbb{D}_{\param} \rnd{\vparam \| \vparam_*} = \breg_{\mathcal{R}} (\vparam \| \vparam_*)=\half \delta \| \vparam - \vparam_*\|^2.\]
We set memory $\memory = \inspace$, where $\inspace$ is the set of all inputs from $\data$. We regularize each example using separate divergences whose Bregman function is equal to the log-partition $A(f)$ (defined in \cref{eq:loss_func}),
\[
   \mathbb{D}_f\rnd{\vf(\vparam)\| \vf(\vparam_*)} = \sum_{i\in\inspace} \breg_{A}(\modelout^i\| \basemodel^i ) = \sum_{i\in\inspace} \loss\rnd{  h(\basemodel^i), h(\modelout^i) }  + \text{constant}.
\]
Smaller memories are discussed later in this section. 
Setting $\tau=1$, we get the following K-prior,
\begin{align}
   \kprior(\vparam; \vparam_*, \inspace) &= \sum_{i\in\inspace} \loss\rnd{  h(\basemodel^i), h(\modelout^i) } + \half \delta \| \vparam - \vparam_*\|^2,
   \label{eq:kprior_glm}
\end{align}
which has a similar form to \cref{eq:GLMobjective}, but the outputs $y_i$ are now replaced by the predictions $h(\basemodel^i)$, and the base model $\vparam_*$ serves as the mean of a Gaussian weight prior. 

We can now show that the gradient of the above K-prior is equal to that of the objective used in \cref{eq:GLMobjective},
\begin{align}
   \nabla \kprior(\vparam; \vparam_*, \inspace) &= \sum_{i\in\inspace} \vphi_i \sqr{  h(\modelout^i) - h(\basemodel^i) } + \delta (\vparam - \vparam_*),  \label{eq:grad_mimic0} \\
    &= \underbrace{ \sum_{i\in\data} \vphi_i \sqr{  h(\modelout^i) - y_i} + \delta\vparam }_{= \nabla \barloss(\text{\vparam}). } \,\, - \,\, \underbrace{ \cancel{ \sum_{i\in\data} \vphi_i \sqr{  h(\basemodel^i) -y_i }  - \delta \vparam_* }}_{= 0.}, 
    \label{eq:grad_mimic}
\end{align}
where the first line is obtained by using \cref{eq:grad_loss} and noting that $\nabla \modelout^i = \vphi_i$, and the second line is obtained by adding and subtracting outputs $y_i$ in the first term.
The second term there is equal to 0 because $\vparam_*$ is a minimizer and therefore $\nabla \barloss(\text{\vparam}_*)=0$. In this case, the K-prior with $\memory=\inspace$ exactly recovers the gradient of the past training objective. 

Why are we able to recover exact gradients? This is because the structure of the K-prior closely follows the structure of \cref{eq:GLMobjective}: the gradient of each term in \cref{eq:GLMobjective} is recovered by a corresponding divergence in the K-prior.
The gradient recovery is due to the property that the gradient of a Bregman divergence is the difference between the \emph{dual} parameters $\nabla\psi(p)$: 
\[ \nabla_{p_1} \mathcal{B}(p_1,p_2) = \nabla\psi(p_1) - \nabla\psi(p_2).\]
This leads to \cref{eq:grad_mimic0}. For the function-space divergence term, $h(\modelout^i) - h(\basemodel^i)$ are the differences in the (dual) expectation parameters. 
For the weight-space divergence term, we note that the dual space is equal to the original parameter space $\mathcal{W}$ for the $L_2$ regularizer, leading to $\vparam - \vparam_*$. 
Lastly, we find that terms cancel out by using the optimality of $\vparam_*$, giving us the exact gradients.

\subsection{K-priors with limited memory}
In practice, setting $\memory = \inspace$ might be as slow as full retraining, but for incremental changes, we may not need all of them (see \cref{fig:fig1}, for example).
Then, which inputs should we include? The answer lies in the gradient-reconstruction error $\ve$, shown below for $\memory \subset \inspace$,
\begin{align}
   \ve(\vparam; \memory) = \nabla \barloss(\vparam) - \nabla \kprior(\vparam; \vparam_*, \memory) &= \sum_{i\in\inspace \backslash \memory} \vphi_i \sqr{  h(\modelout^i) - h(\basemodel^i) }. 
   \label{eq:grad_err}
\end{align}
The error depends on the ``leftover'' $\vphi_i$ for $i \in \inspace \backslash \memory$, and their discrepancies $h(\modelout^i) - h(\basemodel^i)$.
A simple idea could be to include the inputs where predictions disagree the most, but this is not feasible because the candidates $\vparam$ are not known beforehand. The following approximation is more practical,
\begin{equation}
   \ve(\vparam; \memory)  
   \approx  \vG_*(\inspace\backslash\memory) (\vparam - \vparam_*), \textrm{ where } \vG_*(\inspace\backslash\memory) =  \sum_{i\in\inspace\backslash\memory} \vphi_i  h'(\basemodel^i) \vphi_i^\top.
   \label{eq:1order}
\end{equation}
This is obtained by using the Taylor approximation $h(\modelout^i) -  h(\basemodel^i) \approx h'(\basemodel^i) (\nabla \basemodel^i)^\top (\vparam - \vparam_*)$ in \cref{eq:grad_err} ($h'(f^i)$ is the derivative). 
The approximation is conveniently expressed in terms of the Generalized Gauss-Newton (GGN) matrix \citep{martens2014new}, denoted by $\vG_*(\cdot)$.
The approximation suggests that $\memory$ should be chosen to keep the \emph{leftover} GGN matrix $\vG_*(\inspace\backslash\memory)$ orthogonal to $\vparam -\vparam_*$.
Since $\vparam$ changes during training, a reasonable approximation is to choose examples that keep the top-eigenvalues of the GGN matrix. 
This can be done by forming a low-rank approximation by using sketching methods, such as the leverage score \citep{cook1977detection, alaoui2014fast, cohen2015ridge, calandriello2019gaussian}.
A cheaper alternative is to choose the examples with highest $h'(\basemodel^i)$. The quantity is cheap to compute in general, for example, for deep networks it is obtained with just a forward pass.  
Such a set has been referred to as the `memorable past' by \citet{pan2020continual}, who found it to work well for classification.
Due to its simplicity, we will use this method in our experiments, and leave the application of sketching methods as future work.

K-priors with limited memory can achieve low reconstruction error. This is due to an important feature of K-priors: they do not restrict inputs $\vu_i$ to lie within the training set $\inspace$. The inputs can be arbitrary locations in the input space.
This still works because a ground-truth label is not needed for $\vu_i$, and only model predictions $\modelout(\vu_i)$ are used in K-priors. As long as the chosen $\vu_i$ represent $\inspace$ well, we can achieve a low gradient-reconstruction error, and sometimes even perfect reconstruction.
Theoretical results regarding this point are discussed in \cref{app:kprior_svd}, where we present the \emph{optimal} K-prior which can theoretically achieve perfect reconstruction error by using singular vectors of $\vPhi^\top = [\vphi(\vx_1),\vphi(\vx_2), \ldots, \vphi(\vx_N)]$.
When only top-$M$ singular vectors are chosen, the error grows according to the leftover singular values.
The optimal K-prior is difficult to realize in practice, but the result shows that it is theoretically possible to achieve low error with limited memory.

\subsection{Adaptation using K-priors}
We now discuss how the K-prior of \cref{eq:kprior_glm} can be used for the Add/Remove Data tasks.
Other adaptation tasks and corresponding K-priors are discussed in \cref{app:ex_adapt_kpriors}.

Because K-priors can reconstruct the gradient of $\barloss(\vparam)$, we can use them to adapt instead of retraining from scratch. For example, to add/remove data from the GLM solution in \cref{eq:GLMobjective}, we can use the following K-prior regularized objectives,
\begin{align}
   \hat{\vparam}_{+} &= \argmin_{\text{\vparam}\in\mathcal{W}} \,\, \loss_j(\vparam) + \kprior(\vparam; \vparam_*, \memory), 
   &&\quad \hat{\vparam}_{-} = \argmin_{\text{\vparam}\in\mathcal{W}} \,\, - \loss_k(\vparam) + \kprior(\vparam; \vparam_*, \memory) .
   \label{eq:add_kprior}
\end{align}
Using \cref{eq:grad_mimic}, it is easy to show that this recovers the exact solution when all the past data is used. 
\cref{app:ex_adapt_kpriors} details analogous results for the Change Regularizer and Change Model Class tasks.
\begin{thm}
    For $\memory=\inspace$, we have $\vparam_{+} = \hat{\vparam}_{+}$ and $\vparam_{-} = \hat{\vparam}_{-}$.
   \label{thm:equivalence_glm}
\end{thm}
For limited memory, we expect the solutions to be close when memory is large enough. This is because the error in the gradient is given by \cref{eq:grad_err}. The error can be reduced by choosing better $\memory$ and/or by increasing its size to ultimately get perfect recovery.

We stress that \cref{eq:add_kprior} is fundamentally different from replay methods that optimize an objective similar to \cref{eq:addremove} but use a small memory of past examples \cite{robins1995catastrophic}.
Unlike such methods, we use the predictions $h(\basemodel^i)$, which we can think of as soft labels, with potentially more information than the true one-hot encoded labels $y_i$.
Given a fixed memory budget, we expect K-prior regularization to perform better than such replay methods, and we observe this empirically in \cref{sec:experiments}.

\subsection{K-priors for Generic Learning Problems}
The main principle behind the design of K-priors is to construct it such that the gradients can faithfully be reconstructed. 
As discussed earlier, this is often possible by exploiting the structure of the learning problem. For example, to replace an old objective such as \cref{eq:ERMold}, with loss $\loss_i^{\text{old}}(f)$ and regularizer $\oldreg^{\text{old}}(\vparam)$, with a
new objective with loss $\loss_i^{\text{new}}(f)$ and regularizer $\oldreg^{\text{new}}(\vparam)$, the divergences should be chosen such that they have the following gradients,
\begin{equation}
   \nabla \mathbb{D}_\param (\vparam \|\vparam_*) = \nabla \oldreg^{\text{new}}(\vparam) - \nabla \oldreg^{\text{old}}(\vparam),\quad \quad  \nabla \mathbb{D}_f (\vf(\vparam) \| \vf(\vparam_*)) = \nabla \vf(\vparam)^\top \vB \, \vd_m
   \label{eq:genericKpriors}
\end{equation}
where $\vd_m$ is an $M$-length vector with the discrepancy $\nabla \loss^{\text{new}}_i(\modelout^i) - \nabla \loss^{\text{old}}_i(\basemodel^i)$ as the $i$'th entry. The matrix $\vB$ is added to counter the mismatch between $\data$ and $\memory$. Similar constructions can be used for other learning objectives. For non-differentiable functions, a Bayesian version can be used with the Kullback-Leibler (KL) divergence (we discuss
an example in the next section). 
We can use exponential-family distributions which implies a Bregman divergence through KL~\cite{banerjee2005clustering}. 
Since the gradient of such divergences is equal to the difference in the dual-parameters, the general principle is to use divergences with an appropriate dual space to swap the old information with new information. 

\section{K-priors: Extensions and Connections}
\label{sec:unify}

The general principle of adaptation used in K-priors connects many existing works in specific settings. We will now discuss these connections to show that K-priors provide a unifying and generalizing principle for these seemingly unrelated methods in fields such as online learning, deep learning, SVMs, Bayesian Learning, Gaussian Processes, and continual learning. 

\subsection{Weight-Priors}
Quadratic or Gaussian weight-priors \cite{cesa2006prediction, kirkpatrick2017overcoming, schwarz2018progress} be seen as as specialized cases of K-priors, where restrictive approximations are used. For example, the following quadratic regularizer, 
\[ \oldreg_{\text{quad}}(\vparam; \vparam_*) = (\vparam - \vparam_*)^\top \sqr{ \vG_*(\inspace) + \delta\vI } (\vparam - \vparam_*),\]
which is often used in used in online and continual learning \cite{cesa2006prediction, kirkpatrick2017overcoming},
can be seen as a first-order approximation of the K-prior in \cref{eq:kprior_glm}. 
This follows by approximating the K-prior gradient in \cref{eq:kprior_glm} by using the Taylor approximation used in \cref{eq:1order}, to get
\begin{align*}
   \nabla \kprior(\vparam; \vparam_*, \inspace) &\approx \sum_{i\in\inspace} \vphi_i \sqr{ h'(\vphi_i^\top\vparam_*) \vphi_i^\top (\vparam - \vparam_*) } + \delta (\vparam - \vparam_*) \,\, = \nabla \oldreg_{\text{quad}}(\vparam; \vparam_*). 
\end{align*}
K-priors can be more accurate than weight priors but may require larger storage for the memory points. However, we expect the memory requirements to grow according to the rank of the feature matrix (see \cref{app:kprior_svd}) which may still be manageable. If not, we can apply sketching methods.

\subsection{K-priors for Deep Learning and Connections to Knowledge Distillation}
We now discuss the application to deep learning. It is clear that the functional term in K-priors is similar to Knowledge distillation (KD)~\citep{hinton2015distilling}, which is a popular approach for model compression in classification problems using a softmax function (the following expression is from \citet{lopez2015unifying}, also see \cref{app:expts}), 
\begin{equation}
\begin{split}
   \loss_{\text{KD}}(\vparam) = \lambda \sum_{i\in\data} \loss\rnd{ y_i, h(\modelout^i) }+ (1-\lambda) \sum_{i\in\data} \ell\rnd{ h(\basemodel^i/T),\, h(\modelout^i) },
   \label{eq:KD}
\end{split}
\end{equation}
The base model predictions are often scaled with a temperature parameter $T>0$, and $\lambda\in[0,1]$. KD can be seen as a special case of K-priors without the weight-space term ($\tau=0$). K-priors extend KD in three ways, by (i) adding the weight-space term,
(ii) allowing general link functions or divergence functions, and (iii) using a potentially small number of examples in $\memory$ instead of the whole dataset. With these extensions, K-priors can handle adaptation tasks other than compression. Due to their similarity, it is also possible to borrow tricks used in KD to improve the performance of K-priors.

KD often yields solutions that are better than retraining from scratch. Theoretically the reasons behind this are not understood well, but we can view KD as a mechanism to reconstruct the past gradients, similarly to K-priors. As we now show, this gives a possible explanation behind KD's success. Unlike GLMs, K-priors for deep learning do not recover the exact gradient of the past training objective, and there is an additional left-over term (a derivation is in \cref{app:kprior_dl}),
\begin{align}
   \nabla \kprior (\vparam) &= \underbrace{ \sum_{i\in\data} \nabla \modelout^i \sqr{  h(\modelout^i) - y_i} + \delta\vparam }_{ = \nabla \barloss(\text{\vparam}) }  
   - \underbrace{ \rnd{ \sum_{i\in\data} \nabla \modelout^i r_{\text{\vparam}_*}^i  + \delta \vparam_* } }_{\text{Additional term since } \nabla \modelout^i \ne \nabla \basemodel^i}, 
\end{align}
where $r_{\text{\vparam}_*}^i := h(\basemodel^i) - y_i$ is the residual of the base model.
It turns out that the gradient of the KD objective in \cref{eq:KD} has this exact same form  when $\delta=0$, $T=1$ (derivation in \cref{app:kprior_dl}),
      \[ \nabla \loss_{\text{KD}} (\vparam)  = \sum_{i\in\data} \nabla \modelout^i \sqr{  h(\modelout^i) - y_i}   
   - (1-\lambda) \sum_{i\in\data} \nabla \modelout^i r_{\text{\vparam}_*}^i.  \]
The additional term adds large gradients to push away from the high residual examples (the examples the teacher did not fit well). This is similar to Similarity-Control for SVMs from \citet{vapnik2015learning}, where ``slack''-variables are used in a dual formulation to improve the student, who could now be solving a simpler separable classification problem.
The residuals $r^i_{\text{\vparam}_*}$ above play a similar role as the slack variables, but they do not require a dual formulation.
Instead, they arise due to the K-prior regularization in a primal formulation. In this sense, K-priors can be seen as an easy-to-implement scheme for Similarity Control, that could potentially be useful for student-teacher learning.

\citet{lopez2015unifying} use this idea further to generalize distillation and interpret residuals from the teachers as corrections for the student (see Eq. 6~in their paper).
In general, it is desirable to trust the knowledge of the base model and use it to improve the adapted model. These previous ideas are now unified in K-priors: we can provide the information about the decision boundary to the student in a more accessible form than the original data (with true labels) could. 

\subsection{Adding/removing data for SVMs}
K-prior regularized training yields equivalent solutions to the adaptation strategies used in SVM to add/remove data examples. K-priors can be shown to be equivalent to the primal formulation of such strategies~\cite{Liang_Li_2009}. The key trick to show the equivalence is to use the representer theorem which we will now illustrate for the `Add Data' task in \cref{eq:add_kprior}.
Let $\vPhi_+$ be the $(N+1)\times P$ feature matrix obtained on the dataset $\data\cup j$, then by the representer theorem we know that there exists a $\vbeta \in\real^{N+1}$ such that $\vparam_{+} = \vPhi_+^\top \vbeta$. Taking the gradient of \cref{eq:add_kprior}, and multiplying by $\vPhi_+$, we can write the optimality condition as,
\begin{equation}
   0 = \vPhi_+^\top \nabla [\loss_j(\vparam_{+}) + \kprior (\vparam_{+})] = \sum_{i\in \data\cup j} \rnd{ \left. \nabla_{f} \loss(y_i, h(f)) \right\vert_{f = \text{\vbeta}_i^\top \text{\vk}_{i,+} } } \vk_{i,+}  + \delta \vK_+\vbeta ,
   \label{eq:opt_cond_svm}
\end{equation}
 where $\vK_+ = \vPhi_+ \vPhi_+^\top$ and its $i$'th column is denoted by $\vk_{i,+}$.
This is exactly the gradient of the primal objective in the function-space defined over the full batch $\data\cup j$; see Equation 3.6~in \citet{Chapelle_2007}. The primal strategy is equivalent to the more common dual formulations \citep{cauwenberghs2001incremental, tveit2003incremental, Karasuyama_Takeuchi_2010, duan2007decremental, romero2007incremental, laskov2006incremental, Tsai_Lin_Lin_2014}.
The function-space formulations could be computationally expensive, but speed-ups can be obtained by using support vectors. This is similar to the idea of using limited memory in K-priors in \cref{sec:kpriors}.

\subsection{K-priors for Bayesian Learning and Connections to GPs}
K-priors can be seamlessly used for adaptation within a Bayesian learning framework. Consider a Gaussian approximation $q_*(\vparam)$ trained on a variational counterpart of \cref{eq:ERMold} with prior $p(\vparam) \propto \exp\sqr{-\oldreg(\vparam)}$, and its adapted version where we add data, as shown below ($\dkls{}{\cdot}{\cdot}$ is the KL divergence),
\[ q_*(\vparam) = \argmin_{q\in Q} \sum_{i\in\data} \myexpect_q [\barloss_i(\vparam)] + \dkls{}{p}{q}, \,\,\,\, q_+(\vparam) = \argmin_{q\in Q} \sum_{i\in\data\cup j} \myexpect_q [\loss_i(\vparam)] + \dkls{}{p}{q}.\]
Assuming the same setup as \cref{sec:kpriors}, we can recover $q_+(\vparam)$ by using $q_\kprior(\vparam) \propto \exp\sqr{-\kprior(\vparam)}$ where we use the K-prior defined in \cref{eq:kprior_glm} (note that normalization constant of $q_\kprior$ is not required),
\begin{align}
   \hat{q}_{+}(\vparam) = \arg\min_{q\in Q} \,\, \myexpect_{q} [ \loss_j(\vparam)] + \dkls{}{q}{q_\kprior},   
   \label{eq:vi_kpriors}
\end{align}
This follows using \cref{eq:grad_mimic}. Details are in \cref{app:bayes_adapt}.
In fact, when this Bayesian extension is written in the function-space similarly to \cref{eq:opt_cond_svm}, it is related to the online updates used in GPs~\citep{csato2002sparse}. When $q_\kprior$ is built with limited memory, as described in \cref{sec:kpriors}, the application is similar to sparse variational GPs, but now data examples are used as inducing inputs. These connections are discussed in more detail in \cref{app:bayes_adapt}. Our K-prior formulations operates in the weight-space and can be easily trained with first-order methods, however an equivalent formulation in the function space can also be employed, as is clear from these connections.
The above extensions can be extended to handle arbitrary exponential-family approximations by appropriately defining K-priors using KL divergences. We omit these details since this topic is more suitable for a Bayesian version of this paper.

\subsection{Memory-Based Methods for Deep Continual Learning}
K-priors is closely related to recent functional regularization approaches proposed for deep continual learning \cite{li2017learning, rebuffi2017icarl, benjamin2018measuring, schwarz2018progress, titsias2020functional, pan2020continual, buzzega2020dark}. The recent FROMP approach of \citet{pan2020continual} is closest to ours where the form of the functional divergence used is similar to our suggestion in
\cref{eq:genericKpriors}. Specifically, comparing with \cref{eq:genericKpriors}, their functional divergence correspond to the vector $\vd_m$ with the $i$'th entry as $h(f_{\text{\vparam}}(\vu_i)) - h(f_{\text{\vparam}_*}(\vu_i))$ for $\vu_i\in\memory$, and the matrix $\vB$ is (can be seen as Nystrom approximation), 
\[
   \vB = \vLambda(\vparam) \sqr{\vLambda(\vparam_*) \nabla \vf(\vparam_*) \vG(\vparam_*)^{-1} \nabla \vf(\vparam_*)^\top \vLambda(\vparam_*)}^{-1},
\]
where $\vLambda(\vparam)$ is a diagonal matrix with $h'(\modelout(\vu_i))$ as the $i$'th diagonal entry, and $\vG(\vparam_*) = \nabla \vf(\vparam_*)^\top \vLambda(\vparam_*)\nabla \vf(\vparam_*) + \delta \vI$ is the GGN matrix. 
They also propose to use `memorable past' examples obtained by sorting $h'(\basemodel^i)$, which is consistent with our theory (see \cref{eq:grad_err}). Based on our work, we can interpret the approach of \citet{pan2020continual} as a mechanism to reconstruct the gradient of the past, which gives very good performance in practice.

Another related approach is the gradient episodic memory (GEM) \cite{lopez2017gradient}, where the goal is to ensure that the $\sum_{i\in\memory} [\loss(y_i, \basemodel^i)
- \loss(y_i, \modelout^i)] <0$. This is similar in spirit to the student-teacher transfer of \citet{vapnik2015learning} where the loss of the student is regularized using the model output of the teacher (see Eq. 7~in \citet{vapnik2015learning} for an example). \citet{lopez2017gradient} relax the optimization problem to write it in terms of the gradients, which is similar to K-priors, except that K-priors use a first-order optimization method, which is simpler than the dual approach used in
\citet{lopez2017gradient}.

Most of these approaches do not employ a weight-space divergence, and sometimes even the function-space divergence is replaced by the Euclidean one \citep{benjamin2018measuring, buzzega2020dark}. Often, the input locations are sampled randomly, or using a simple replay method \cite{buzzega2020dark} which could be suboptimal. 
Some approaches propose computationally-expensive methods for choosing examples to store in memory \citep{aljundi2019online, aljundi2019gradient}, and some can be seen as related to choosing points with high leverage \citep{aljundi2019gradient}.
The approach in \citet{titsias2020functional} uses inducing inputs which is closely connected to the online GP update. The method we proposed does not contradict with these, but gives a more direct way to choose the points where the gradient errors are taken into consideration.

\section{Experimental Results}
\label{sec:experiments}

We compare the performance of K-priors to retraining with full-batch data (`Batch') and a retraining with replay from a small memory (`Replay'), and use $\tau=1$. For fair comparisons, we use the same memory for Replay and K-priors obtained by choosing points with highest $h'(\basemodel^i)$ (see \cref{sec:kpriors}).  Memory chosen randomly often gives much worse results and we omit these results. Replay uses the true label while K-priors use model-predictions. We compare these three methods on the four adaptation tasks: `Add Data', `Remove Data', `Change Regularizer', and `Change
Architecture'. For the `Add Data' task, we also compare to Weight-Priors with GGN. 

\begin{figure}[t]
\centering
     \includegraphics[width=\textwidth]{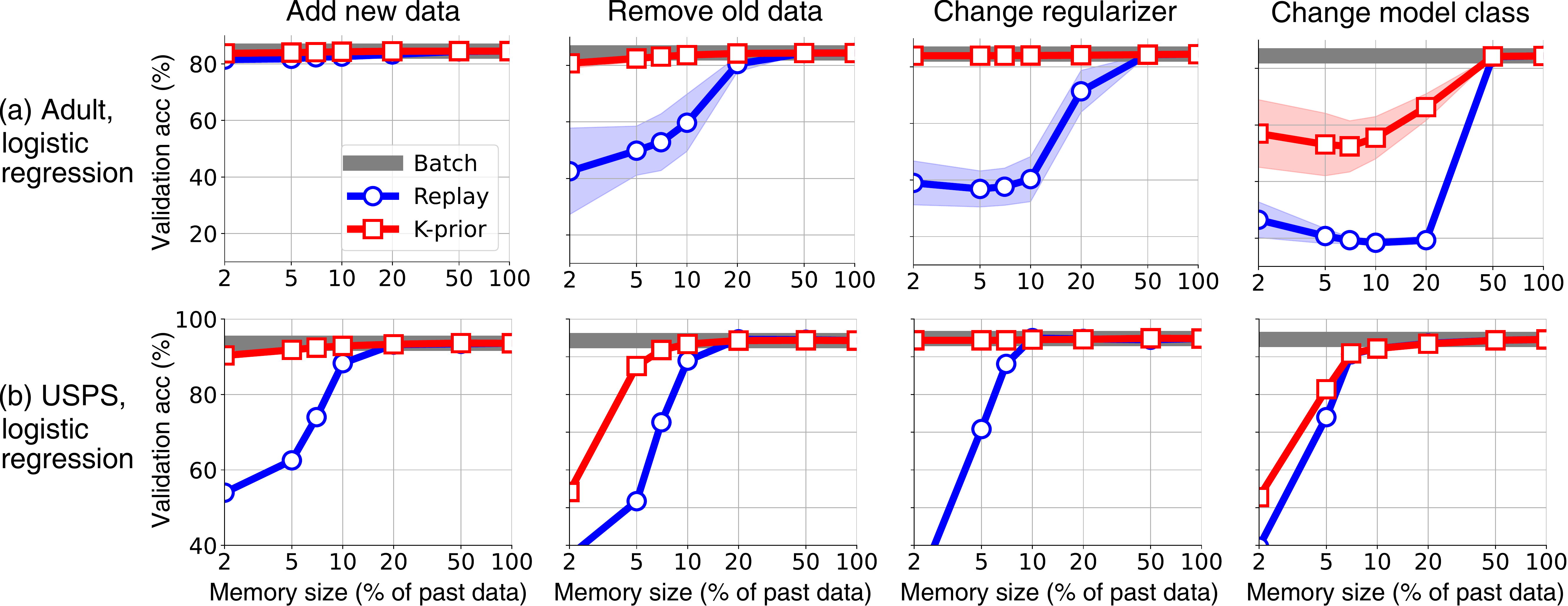}
     \caption{K-priors (red squares) match Batch (grey) while mostly using 2-5\% of the data (for only 3 tasks a larger fraction is required). K-priors always outperforms Replay which uses the true labels. K-priors replace the labels by the model predictions (see the discussion after \cref{thm:equivalence_glm}).}
    \label{fig:changing_mem_plots}
\end{figure}

\begin{figure}[t]
\centering
    \includegraphics[width=\textwidth]{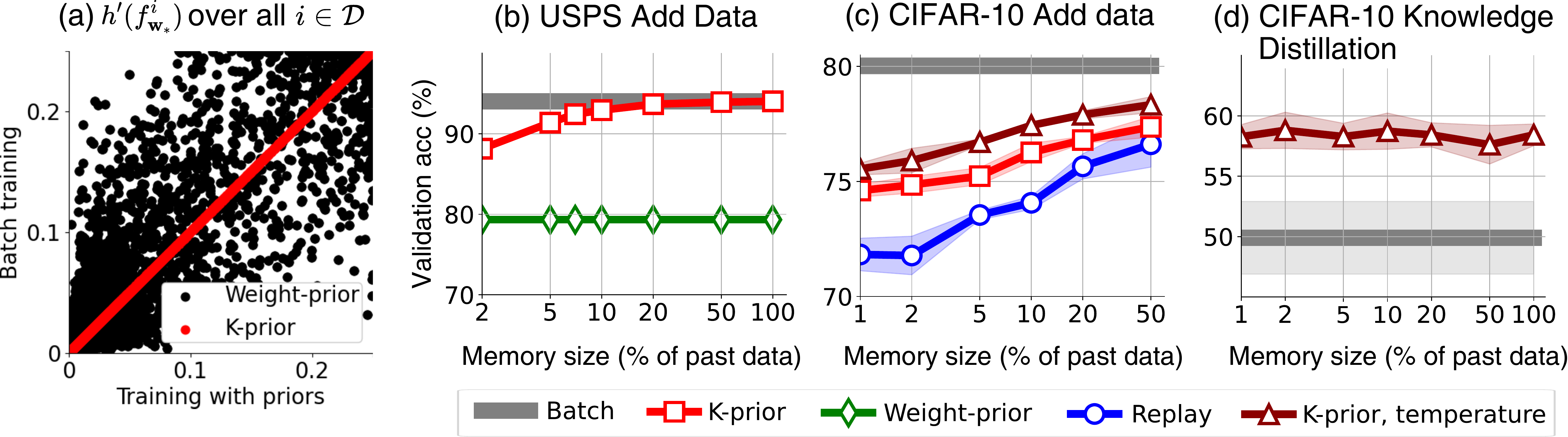}
    \caption{(a) When compared at the Batch solution for the `Add Data' task on USPS, weight priors give incorrect values of $h'(\modelout^i)$ (shown with black dots, each dot correspond to a data examples).  Points on the diagonal means a perfect match which is the case for K-priors (show with red dots).
    (b) Due to this, weight-priors (green diamonds) perform worse than K-priors (red squares). 
    (c) For the 'Add data' task on CIFAR-10, K-priors outperform replay (blue circles), but performance can still be improved by using a temperature parameter (dark-red triangles).
    (d) The same is true for knowledge distillation \citep{hinton2015distilling}, and we see that we can reduce memory size while still performing better than the student model. 
    }
    \label{fig:fig3}
\end{figure}

Our overall finding is that, for GLMs and deep learning on small problems, K-priors can achieve the same performance as Batch, but with a small fraction of data (often 2-10\% of the full data (\cref{fig:fig1}, right, and \cref{fig:changing_mem_plots}). Replay does much worse for small memory size, which clearly shows the advantage of using the model predictions (instead of true labels) in K-priors. Weight priors generally perform well, but they can do badly when the adaptation
involves a drastic change for examples in $\memory$ (see \cref{fig:fig3}). Finally, for large deep-learning
problems, results are promising but more investigations are required with extensive hyperparameter tuning.

Several additional experiments are in \cref{app:expts}. 
In \cref{app:expt_rand_init}, we study the effect of randomly initialization and find that it performs similarly to a warm start at $\vparam_*$.
In \cref{app:expt_num_backprops}, we find that K-priors with limited memory are take much less time to reach a specified accuracy than both batch and replay. The low cost is due to a small memory size. 
Replay also uses small memory but performs poorly.

\textbf{Logistic Regression on the `UCI Adult' dataset.} This is a binary classification problem consisting of 16,100 examples to predict income of individuals. We randomly sample 10\% of the training data (1610 examples), and report mean and standard deviation over 10 such splits. For training, we use the L-BFGS optimizer for logistic regression with polynomial basis. 
Results are summarized in \cref{fig:changing_mem_plots}(a).
For the `Add Data' task, the base model uses 9\% of the data and we add 1\% new data. For `Remove Data', we remove 100 data examples (6\% of the training set) picked by sorting $h'(\basemodel^i)$). 
For the `Change Regularizer' task, we change the $L_2$ regularizer from $\delta=50$ to $5$, and for `Change Model Class', we reduce the polynomial degree from $2$ to $1$. 
K-priors perform very well on the first three tasks, remaining very close to Batch, even when the memory sizes are down to 2\%. `Changing Model Class' is slightly challenging, but K-priors still significantly out-perform Replay.

\textbf{Logistic Regression on the `USPS odd vs even' dataset.} The USPS dataset consists of 10 classes (one for each digit), and has 7,291 training images of size $16\times16$. We split the digits into two classes: odd and even digits. 
Results are in \cref{fig:changing_mem_plots}(b).
For the `Add Data' task, we add all examples for the digit 9 to the rest of the dataset, and for `Remove Data' we remove the digit 8 from the whole dataset. By adding/removing an entire digit, we enforce an \emph{inhomogeneous} data split, making the tasks more challenging.
The `Change Regularizer' and `Change Model Class' tasks are the same as the Adult dataset.
K-priors perform very well on the `Add Data' and `Change Regularizer' tasks, always achieving close to Batch performance. For `Remove Data', which is a challenging task due to inhomogenity, K-priors still only need to store 5\% of past data to maintain close to 90\% accuracy, whereas Replay requires 10\% of the past data.

\textbf{Neural Networks on the `USPS odd vs even' dataset.} This is a repeat of the previous experiment but with a neural network (a 1-hidden-layer MLP with 100 units). Results are in \cref{fig:fig1} (right).
The `Change Regularizer' task now changes $\delta=5$ to $10$, and the `Change Architecture' task compresses the architecture from a 2-hidden-layer MLP (100 units per layer) to a 1-hidden-layer MLP with 100 units. We see that even with neural networks, K-priors perform very well, similarly out-performing Replay and remaining close to the Batch solution at small memory sizes.

\textbf{Weight-priors vs K-priors.} As discussed in the main text, weight-priors can be seen as an approximation of K-priors where $h'(\modelout^i)$ are replaced by `stale' $h'(\basemodel^i)$, evaluated at the old $\vparam_*$. In \cref{fig:fig3}(a), we visualize these `stale' $h'(\basemodel^i)$ and compare them to K-priors which obtains values close to the ones found by Batch.
Essentially, for the points at the diagonal the match is perfect, and we see that it is the case for K-priors but not for the weight-priors.
We use logistic regression on the USPS data (the `Add Data' task).
This inhomogeneous data split is difficult for weight-priors, and we show in \cref{fig:fig3}(b) that weight-priors do not perform well. 
For homogeneous data splits, weight-priors do better than this, but this result goes to show that they do not have any mechanisms to fix the mistakes made in the past. In K-priors, we can always change $\memory$ to improve the performance. We provide more plots and results in \cref{app:expts}, including results for weight-priors on all the `Add data' tasks we considered in this paper.

\textbf{MNIST and CIFAR-10, neural networks.} 
Finally, we discuss results on larger problems in deep learning. 
We show many adaptation tasks in \cref{app:expts} for 10-way classification on MNIST \cite{lecun-mnisthandwrittendigit-2010} with MLPs and 10-way classification on CIFAR-10 with CifarNet \cite{zenke2017continual}, trained with the Adam optimizer \cite{kingma2014adam}. 
In \cref{fig:fig3}(c) we show one representative result for the `Add data' task with CIFAR-10, where we add a random 10\% of CIFAR-10 training data to the other 90\% (mean and standard deviation over 3 runs). Although vanilla K-priors outperform Replay, there is now a bigger gap between K-prior and Batch even with 50\% past data stored. 
However, when we use a temperature (similar to knowledge distillation in \eqref{eq:KD} but with the weight term included), K-priors improves. 

A similar result is shown in \cref{fig:fig3}(d) for knowledge distillation ($\delta=0$ but with a temperature parameter) where we are distill from a CifarNet teacher to a LeNet5-style student (details in \cref{app:expts}). Here, K-priors with 100\% data is equivalent to Knowledge Distillation, but when we reduce the memory size using our method, we still outperform Batch (which is trained from scratch on all data).
Overall, our initial effort here suggests that K-priors can do better than Replay, and have potential to give better results with more hyperparameter tuning.

\section{Discussion}
In this paper, we proposed a class of new priors, called K-prior. We show general principles of obtaining accurate adaptation with K-priors which are based on accurate gradient reconstructions. The prior applies to a wide-variety of adaptation tasks for a range of models, and helps us to connect many existing, seemingly-unrelated adaptation strategies in ML. Based on our adaptation principles, we
derived practical methods to enable adaptation by tweaking models' predictions at a few past examples. This is analogous to adaptation in humans and animals where past experiences is used for new situations. In practice, the amount of required past memory seems sufficiently low. 

The financial and environmental costs of retraining are a huge concern for ML practitioners, which can be reduced with quick adaptations. The current pipelines and designs are specialized for an offline, static setting. Our approach here pushes towards a simpler design which will support a more dynamic setting. The approach can eventually lead to new systems that learn quickly and flexibly, and also act
sensibly across a wide range of tasks.
This opens a path towards systems that learn incrementally in a continual fashion, with the potential to fundamentally change the way ML is used in scientific and industrial applications.
We hope that this work will help others to do more towards this goal in the future. We ourselves will continue to push this work in that direction.

\section*{Acknowledgements}
We would like to thank the members of the Approximate-Bayesian-Inference team at RIKEN-AIP. Special thanks to Dr. Thomas Möllenhoff (RIKEN-AIP), Dr. Gian Maria Marconi (RIKEN-AIP), Peter Nickl (RIKEN-AIP), and also to Prof. Richard E. Turner (University of Cambridge).
Mohammad Emtiyaz Khan is partially supported by KAKENHI Grant-in-Aid for Scientific Research (B), Research Project Number 20H04247.
Siddharth Swaroop is partially supported by a Microsoft Research EMEA PhD Award.

\section*{Author Contributions Statement}
List of Authors: Mohammad Emtiyaz Khan (M.E.K.), Siddharth Swaroop (S.S.).

Both the authors were involved in the idea conception. S.S. derived a version of theorem 1 with some help from M.E.K. This was then modified and generalized by M.E.K. for generic adaptation tasks. The general principle of adaptation described in the paper are due to M.E.K., who also derived connections to SVMs and GPs, and extensions to Bayesian settings (with regular feedback from S.S.). Both authors worked together on the connections to Knowledge Distillation and Deep Continual Learning.
S.S. performed all the experiments (with regular feedback from M.E.K.). M.E.K. wrote the main sections with the help of S.S., and S.S. wrote the section about the experiments. Both authors proof-read and reviewed the paper.

\bibliographystyle{plainnat}

\bibliography{refs}

\newpage
\appendix
\section{Optimal K-priors for GLMs}
\label{app:kprior_svd}

We present theoretical results to show that K-priors with limited memory can achieve low gradient-reconstruction error. We will discuss the \emph{optimal} K-prior which can theoretically achieve perfect reconstruction error. Note that the prior is difficult to realize in practice since it requires all past training-data inputs $\inspace$. Our goal here is to establish a theoretical limit, not to give practical choices.

Our key idea is to choose a few input locations that provide a good representation of the training-data inputs $\inspace$. 
We will make use of the singular-value decomposition (SVD) of the feature matrix,
\[\vPhi^\top = \vU_{1:K}^* \vS^*_{1:K} (\vV^*_{1:K})^\top\]
where $K\le \min(N,P)$ is the rank, $\vU_{1:K}^*$ is $P\times K$ matrix of left-singular vectors $\vu_i^*$, $\vV_{1:K}^*$ is $N\times K$ matrix of right-singular vectors $\vv_i^*$, and $\vS_{1:K}^*$ is a diagonal matrix with singular values $s_i$ as the $i$'th diagonal entry. 

We define $\memory^* = \{\vu_1^*, \vu_2^*, \ldots, \vu_K^*\}$, and the following K-prior,
\begin{align}
   \kprior_{\text{opt}}(\vparam; \vparam_*, \memory^*) &= \sum_{j=1}^K \beta_j^* \loss\rnd{  h(\basemodel(\vu_j^*)), h(\modelout(\vu_j^*)) } + \half \delta \| \vparam - \vparam_*\|^2.
   \label{eq:kprior_glm_svd}
\end{align}
Here, each functional divergence is weighted by $\vbeta_j^*$ which refers to the elements of the following, 
\[
   \vbeta^* = \vD_u^{-1} \vS_{1:K}^* \vV_{1:K}^\top \vd_x
\]
where $\vd_x$ is an $N$-length vector with entries $h(\modelout^i) - h(\basemodel^i)$ for all $i\in\inspace$, while $\vD_u$ is a $K\times K$ diagonal matrix with diagonal entries $h(\modelout(\vu_j)) - h(\basemodel(\vu_j))$ for all $j = 1,2, \ldots, K$.
The above definition departs slightly from the original definition where only a single $\tau$ is used.
The weights $\beta_j^*$ depend on $\inspace$, so it is difficult to compute them in practice when the memory is limited. However, it might be possible to estimate them for some problems.

Nevertheless, with $\beta_j^*$, the above K-prior can be achieve perfect reconstruction. The proof is very similar to the one given in \cref{eq:grad_mimic0,eq:grad_mimic}, and is shown below,
\begin{align*}
   \nabla \kprior_{\text{opt}}(\vparam; \vparam_*, \memory^*) &= \sum_{j=1}^K \beta_j^* \vu_i^* \sqr{  h(\modelout(\vu_i)) - h(\basemodel(\vu_i)) } + \delta (\vparam - \vparam_*), \\
   &= \vU_{1:K}^* \vD_u \vbeta_*  + \delta (\vparam - \vparam_*), \\
   &= \vU_{1:K}^* \vS_{1:K}^* \vV_{1:K}^\top \vd_x   + \delta (\vparam - \vparam_*), \\
   &= \vPhi^\top \vd_x   + \delta (\vparam - \vparam_*), \\
   &= \sum_{i\in\inspace} \vphi_i \sqr{  h(\modelout^i) - h(\basemodel^i) } + \delta (\vparam - \vparam_*), \\
   &= \underbrace{ \sum_{i\in\data} \vphi_i \sqr{  h(\modelout^i) - y_i} + \delta\vparam }_{= \nabla \barloss(\text{\vparam}). } \,\, - \,\, \underbrace{ \cancel{ \sum_{i\in\data} \vphi_i \sqr{  h(\basemodel^i) -y_i }  - \delta \vparam_* }}_{= 0.}. 
\end{align*}
The first line is simply the gradient, which is then rearranged in the matrix-vector product in the second line. The third line uses the definition of $\vbeta_*$, and the fourth line uses the SVD of $\vPhi$. In the fifth line we expand it to show that it is the same as \cref{eq:grad_mimic0}, and the rest follows as before.

Due to their perfect gradient-reconstruction property, we call the prior in \cref{eq:kprior_glm_svd} the \emph{optimal} prior. When only top-$M$ singular vectors are chosen, the gradient reconstruction error grows according to the leftover singular values. We show this below where we have chosen $\memory_M^* = \{\vu_1^*, \vu_2^*, \ldots, \vu_M^*\}$ as the set of top-$M$ singular vectors, 
\begin{align*}
   \ve_{\text{opt}}(\vparam; \vparam, \memory_M^*) &=  \nabla \barloss(\text{\vparam}) - \nabla \kprior_{\text{opt}}(\vparam; \vparam_*, \memory_M^*) \\
   &= \nabla \barloss(\text{\vparam}) - \nabla \kprior_{\text{opt}}(\vparam; \vparam_*, \memory^*) + \nabla \kprior_{\text{opt}}(\vparam; \vparam_*, \memory^*) - \nabla \kprior_{\text{opt}}(\vparam; \vparam_*, \memory_M^*) \\
   &= \nabla \kprior_{\text{opt}}(\vparam; \vparam_*, \memory^*) - \nabla \kprior_{\text{opt}}(\vparam; \vparam_*, \memory_M^*) \\
   &= \sum_{j=M+1}^M \beta_j^* \vu_i^* \sqr{  h(\modelout(\vu_i)) - h(\basemodel(\vu_i)) }, \\
   &= \vU_{M+1:K}^* \vS_{M+1:K}^* \vV_{M+1:K}^\top \vd_x . 
\end{align*}
The first line is simply the definition of the error, and in the second line we add and subtract the optimal K-prior with memory $\memory^*$. The next few lines use the definition of the optimal K-prior and rearrange terms.

Using the above expression, we find the following error,
\[\|\ve_{\text{opt}}(\vparam; \vparam, \memory_M^*)\| = \sqrt{ \Sigma_{j=M+1}^K s_j^2 (a_j^x)^2 }\]
where $a_j^x$ is the $j$'th entry of a vector $\va = \vV_{1:K}^\top \vd_x$. The error depends on the leftover singular values. The error is likely to be the optimal error achievable by any memory of size $M$, and establishes a theoretical bound on the best possible performance achievable by any K-prior.

\section{Additional Examples of Adaptation with K-priors}
\label{app:ex_adapt_kpriors}

Here, we briefly discuss the K-prior regularization for the other adaptation tasks. 

\subsection{The Change Regularizer task}

For Change Regularizer task, we need to slightly modify the K-prior of \cref{eq:kprior_glm}. We replace the weight-space divergence in \cref{eq:kprior_glm} with a Bregman divergence defined using two different regularizers (see Proposition 5~in \citet{nielsen2021variational}),
\begin{equation}
   \breg_{\newreg\oldreg}(\vparam\|\vparam_*) = \newreg(\vparam) + \oldreg^*(\veta_*) - \vparam^\top\veta_*,
   \label{eq:breg_gr}
\end{equation}
where $\veta_* = \nabla \oldreg(\vparam_*)$  is the dual parameter and $\oldreg^*$ is the convex-conjugate of $\oldreg$. This is very similar to the standard Bregman divergence but uses two different (convex) generating functions.

To get an intuition, consider hyperparameter-tuning for the $L_2$ regularizer $\oldreg(\vparam) = \half \delta\|\vparam\|^2$, where our new regularizer $\newreg(\vparam) = \half \gamma \|\vparam\|^2$ uses a hyperparameter $\gamma \ne \delta$. Since the conjugate
$\oldreg^*(\veta) = \half \|\veta\|^2/\delta$
and $\veta_* = \nabla \oldreg(\vparam_*) = \delta\vw_*$, we get
\[
\breg_{\newreg\oldreg}(\vparam\|\vparam_*) = \half ( \gamma \|\vparam\|^2 +  \delta \|\vparam_*\|^2 - 2 \delta \vparam^\top \vparam_*).
\]
When $\gamma = \delta$, then this reduces
to the divergence used in \cref{eq:kprior_glm}, but otherwise it enables us to reconstruct the gradient of the past objective but with the new regularizer.
We define the following K-prior where the weight-divergence is replaced by \cref{eq:breg_gr}, and use it to obtain $\hat{\vparam}_{\newreg}$,
\begin{equation}
   \begin{split}
      \kprior(\vparam; \vparam_*, \memory) &= \sum_{i\in\memory} \loss\rnd{  h(\basemodel^i), h(\modelout^i) } +  \breg_{\newreg\oldreg}(\vparam\|\vparam_*), \\
      \hat{\vparam}_\newreg &= \argmin_{\text{\vparam}\in\mathcal{W}} \kprior(\vparam; \vparam_*, \memory)  
   \end{split}
   \label{eq:reg_swap}
\end{equation}
The following theorem states the recovery of the exact solution.
\begin{thm}
   For $\memory=\inspace$ and strictly-convex regularizers, we have $\vparam_{\newreg} = \hat{\vparam}_{\newreg}$.
   \label{thm:equivalence_glm_2}
\end{thm}
The derivation is very similar to \cref{eq:grad_mimic}, where $\delta(\vparam -\vparam_*)$ is replaced by $\nabla \newreg(\vparam) - \nabla \oldreg(\vparam_*)$. 

\subsection{The Change Model Class task}
We discuss the `Change Model Class' task through an example. Suppose we want to remove the last feature from $\vphi_i$ so that $\vparam\in\real^P$ is replaced by a smaller weight-vector $\vtheta\in\real^{P-1}$. Assuming no change in the hyperparameter, we can simply use a weighting matrix to `kill' the last element of $\vparam_*$. We define the matrix $\vA = \vI_{P-1\times P}$ whose last column is 0 and the rest is the identity matrix of size $P-1$. With this, we can use the following training procedure over a smaller space $\bar{\vparam}$, 
\begin{align}
   \kprior(\vtheta) = \sum_{i\in\memory} \loss\rnd{  h(\basemodel^i), h(f_{\text{\vtheta}}(\vx_i)) } +  \breg_{\oldreg}(\vtheta\|\vA\vparam_*),
   \quad\quad \hat{\vtheta}_* = \argmin_{\text{\vtheta}\in{\Theta}}
   \kprior (\vtheta) 
\end{align}
If the hyperparameters or regularizer are different for the new problem, then the Bregman divergence shown in \cref{eq:breg_gr} can be used, with an appropriate weighting matrix. 

Model compression is a specific instance of the `Change Model Class' task, where the architecture is entirely changed. For neural networks, this also changes the meaning of the weights and the regularization term may not make sense. In such cases, we can simply use the functional-divergence term in K-priors, 
\begin{equation}
   \kprior(\vtheta) = \sum_{i\in\memory} \loss\rnd{  h(\basemodel^i), h(f_{\text{\vtheta}}(\vx_i)) },
   \quad\quad \hat{\vtheta}_* = \argmin_{\text{\vtheta}\in{\Theta}} \kprior(\vtheta)
\end{equation}
This is equivalent to knowledge distillation (KD) in \cref{eq:KD} with $\lambda = 0$ and $T=1$.

Since KD performs well in practice, it is possible to use a similar strategy to boost K-prior, e.g., we can define the following,
\begin{equation}
   \hat{\vtheta}_* = \argmin_{\text{\vtheta}\in{\Theta}}\,\, \lambda  \sum_{i\in\memory} \loss(y_i, h(f_{\text{\vtheta}}^i)) + (1-\lambda) \kprior(\vtheta)
\end{equation}
We could even use limited-memory in the first term. The term $\lambda$ lets us trade-off teacher predictions with the actual data.

We can construct K-priors to change multiple things at the same time, for example, changing the regularizer, the model class, and adding/removing data. A K-prior for such situations can be constructed using the same principles we have detailed.

\section{Derivation of the K-priors Gradients for Deep Learning}
\label{app:kprior_dl}

The gradient is obtained similarly to \eqref{eq:grad_mimic} where we add and subtract $y_i$ in the first term in the first line below,
\begin{align*}
   \nabla \kprior(\vparam) &= \sum_{i\in\inspace} \nabla \modelout^i \sqr{  h(\modelout^i) - h(\basemodel^i) } + \delta (\vparam - \vparam_*), \\
   &= \underbrace{ \sum_{i\in\data} \nabla \modelout^i \sqr{  h(\modelout^i) - y_i} + \delta\vparam }_{ = \nabla \barloss(\text{\vparam}) }  
   - \underbrace{ \sum_{i\in\data} \nabla \modelout^i [ h(\basemodel^i) - y_i] - \delta \vparam_* }_{\ne \nabla \barloss(\text{\vparam}_*), \text{ because } \nabla \modelout^i \ne \nabla \basemodel^i}, 
\end{align*}
The second term is not zero because $\nabla \modelout^i \ne \nabla \basemodel^i$ to get $\nabla \barloss(\vparam_*)$ in the second term.

The gradient of the KD objective can be obtained in a similar fashion, where we add and subtract $y_i$ in the second term in the first line to get the second line,
\begin{align*}
       \nabla \loss_{\text{KD}} (\vparam)  &= \lambda \sum_{i\in\data} \nabla \modelout^i \sqr{  h(\modelout^i) - y_i}  + (1-\lambda) \sum_{i\in\data} \nabla \modelout^i \sqr{ h(\modelout^i) - h(\basemodel^i) } ,\\ 
       &= \sum_{i\in\data} \nabla \modelout^i \sqr{  h(\modelout^i) - y_i}  - (1-\lambda) \sum_{i\in\data} \nabla \modelout^i \sqr{ h(\basemodel^i) - y_i } .
\end{align*}

\section{Proof for Adaptation for Bayesian Learning with K-priors}
\label{app:bayes_adapt}

To prove the equivalence of \eqref{eq:vi_kpriors} to the full batch variational inference problem with a Gaussian $q(\vparam) = \gauss(\text{\vparam}|\text{\vmu}, \text{\vSigma})$, we can use the following fixed point of the variational objective (see Section 3~in \citep{emti2020bayesprinciple} for the expression), 
\begin{align}
   0 &= \left. \nabla_{\text{\vmu}} \myexpect_q [\mathcal{L}(\vparam)]\ \right\vert_{\text{\vmu} =\text{\vmu}_+, \text{\vSigma} = \text{\vSigma}_+} = \left.\myexpect_q [ \nabla_{\text{\vparam}} \mathcal{L}(\vparam) ] \right\vert_{\text{\vmu}=\text{\vmu}_+, \text{\vSigma} = \text{\vSigma}_+} , \label{eq:gp_stat1}\\
   \vSigma_+^{-1} &= \left. \nabla_{\text{\vSigma}} \myexpect_q [ \mathcal{L}(\vparam)] \right\vert_{\text{\vmu}=\text{\vmu}_+, \text{\vSigma} = \text{\vSigma}_+} = \left.\myexpect_q [ \nabla^2_{\text{\vparam}} \mathcal{L}(\vparam) ] \right\vert_{\text{\vmu}=\text{\vmu}_+, \text{\vSigma} = \text{\vSigma}_+}\label{eq:gp_stat2}, 
\end{align}
where $\mathcal{L}(\vparam) = [ \loss_j(\vparam) + \barloss(\vparam) + \oldreg(\vparam)]$, $\vmu_+$ and $\vSigma_+$ are the mean and covariance of the optimal $q_+(\vparam)$ for the `Add Data' task.
For GLMs, both the gradient and Hessian of $\barloss(\vparam)$ is equal to those of $\kprior(\vparam)$ defined in \eqref{eq:kprior_glm}, which proves the equivalence.

For equivalence to GPs, we first note that, similarly to the representer theorem, the mean and covariance of $q_+(\vparam)$ can be expressed in terms of the two $N$-length vectors $\valpha$ and $\vlambda$ \citep{Opper:09, Khan14nips, KhanAFS13},
\[
   \vmu_+ = \vPhi_+^\top \valpha, \quad\quad 
   \vSigma_+ = (\vPhi_+^\top \vLambda \vPhi_+ + \delta\vI)^{-1}, 
\]
where $\vLambda$ is a diagonal matrix with $\vlambda$ as the diagonal. Using this, we can define a marginal $q(f_i) = \gauss(f_i|m_i,v_i)$, where $f_i = \vphi_i^\top \vparam$, with the mean and variance defined as follows,
\[
   m_i = \vphi_i^\top \vmu_+ = \vk_{i,+}^\top \valpha, \quad\quad 
   v_i = \vphi_i^\top \vSigma_+\vphi_i = k_{ii,+} - \vk_{i,+}^\top\rnd{ \vLambda^{-1} + \delta \vK_+}^{-1} \vk_{i,+}, \quad\quad 
\]
where $k_{ii,+} = \vphi_i^\top \vphi_i$. Using these, we can now rewrite the optimality conditions in the function-space to show equivalence to GPs.

We show this for the first optimality condition \eqref{eq:gp_stat1}, 
\begin{align*}
   \left. \nabla_{\text{\vmu}} \myexpect_q [ \mathcal{L}(\vparam)] \right\vert_{\text{\vmu} =\text{\vmu}_+, \text{\vSigma} = \text{\vSigma}_+} 
   = \hspace{-0.5em} \sum_{i\in\data\cup j} \myexpect_{\text{\gauss}(\epsilon_i|0, 1)} \sqr{ \nabla_{f} \loss(y_i, h(f)) \vert_{f= \text{\vphi}_i^\top\text{\vmu}_+ +  \rnd{\text{\vphi}_i^\top\text{\vSigma}_+\text{\vphi}_i}^{1/2}\epsilon_i} } \vphi_i + \delta \vmu_+
\end{align*}
Multiplying it by $\vPhi_+$, we can rewrite the gradient in the function space,
\begin{align*}
   0 = \sum_{i\in\data\cup j} \myexpect_{\text{\gauss}(\epsilon_i|0, 1)} &\sqr{ \nabla_{f} \loss(y_i, h(f)) \vert_{f= m_i +  v_i^{1/2}\epsilon_i} } \vk_{i,+} + \delta \vK_+\valpha \\
   &= \sum_{i\in\data\cup j}  \nabla_{m_i} \myexpect_{q(f_i)} \sqr{ \loss(y_i, h(f_i)) } \vk_{i,+} + \delta \vK_+\valpha
\end{align*}
where $\vm$ is the vector of $m_i$. Setting this to 0, gives us the first-order condition for a GP with respect to the mean, e.g., see Equation 3.6 and 4.1~in \citet{Chapelle_2007}. It is easy to check this for GP regression, where $\loss(y_i, h(f_i)) = (y_i - f_i)^2$, in which case, the equation becomes,
\[
   0 = \sum_{i\in\data\cup j}  (m_i-y_i) \vk_{i,+} + \delta \vK_+\valpha \quad \Rightarrow \valpha = (\vK_+ + \delta\vI)^{-1}\vy,
\]
which is the quantity which gives us the posterior mean. A similar condition condition for the covariance can be written as well.

Clearly, when we use a limited memory, some of the data examples are removed and we get a sparse approximation similarly to approaches such as informative vector machine which uses a subset of data to build a sparse approximation \citep{NIPS2002_d4dd111a}. Better sparse approximations can be built by carefully designing the functional divergence term. For example, we can choose the matrix $\vB$ in the divergence, 
\[
  \mathbb{D}_f (\vf(\vparam) \| \vf(\vparam_*)) = \half\vd_m^\top \vB \vd_m \quad\quad \Rightarrow \quad \nabla \mathbb{D}_f (\vf(\vparam) \| \vf(\vparam_*)) = \nabla \vf(\vparam)^\top \vB \vd_m
\]
This type of divergence is used in \citet{pan2020continual}, where the matrix $\vB$ is set to correlate the examples in $\memory$ with the examples in $\data$. Design of such divergence function is a topic which requires more investigation in the future.

\section{Further experimental results}
\label{app:expts}

We provide more details on all our experiments, such as hyperparameters and more results.

\subsection{Adaptation tasks}
\label{app:expts_adaptation}

\textbf{Logistic Regression on the `UCI Adult' dataset.}
In \cref{fig:changing_mem_plots}(a)  we show results for the 4 adaptation tasks on the UCI Adult dataset, and provide experimental details in \cref{sec:experiments}. 
Note that for all but the `Change Model Class' task, we used polynomial degree 1. 
For all but the `Change Regularizer' task, we use $\delta=5$.

\begin{figure}[ht]
\centering
    \includegraphics[width=0.5\textwidth]{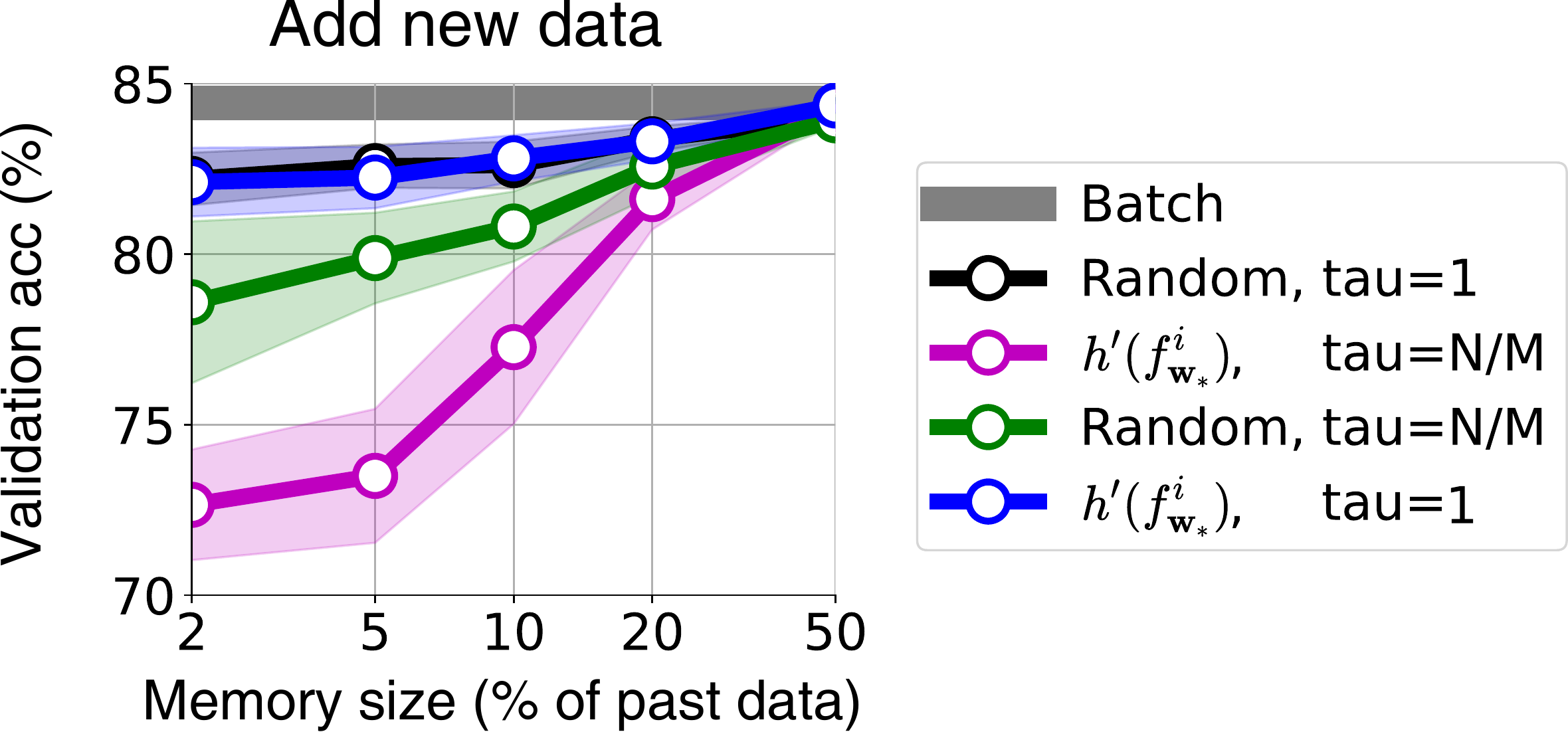}
    \caption{This figure shows using $\tau=1$ works well for Replay, both for random selection of memory and choosing memory by sorting $h'(\basemodel^i)$.
    We compare different methods for Replay on the Adult `Add Data' task. `Random' means the points in memory are chosen randomly as opposed to choosing the points with highest $h'(\basemodel^i)$. We also consider using $\tau=N/M$ instead of $\tau=1$. 
    Choosing randomly or by $h'(\basemodel^i)$ are within standard deviations in this task, so we choose to report memory chosen by $h'(\basemodel^i)$ in other experiments (this is then consistent with the memory in K-priors).}
    \label{fig:adult_replay_different}
\end{figure}

We optimize using LBFGS (default PyTorch implementation) with a learning rate of $0.01$ until convergence.
Throughout our experiments in the paper, we used the same memorable points for Replay as for K-priors (the points with the highest $h'(\basemodel^i)$), and used $\tau=1$ (from \cref{eq:kpost_def}). 
In \cref{fig:adult_replay_different} we provide an ablation study for Replay with different strategies: (i) we choose points by $h'(\basemodel^i)$ and use $\tau=N/M$, (ii) we choose points randomly and use $\tau=1$, (iii) we choose points randomly and use $\tau=N/M$. Recall that $N$ is the past data size (the size of $\data$) and $M$ is the number of datapoints stored in memory (the size of $\memory$). 
We see that choosing points by $h'(\basemodel^i)$ and using $\tau=1$ performs very well, and we therefore choose this for all our experiments.

\textbf{Logistic Regression on the `USPS odd vs even' dataset.}
For all but the `Change Model Class' task, we used polynomial degree 1. 
For all but the `Change Regularizer' task, we use $\delta=50$.
We optimize using LBFGS with a learning rate of $0.1$ until convergence.

\textbf{Neural Networks on the `USPS odd vs even' dataset.}
For all but the `Change Regularizer' task, we use $\delta=5$.
We optimize using Adam with a learning rate of $0.005$ for 1000 epochs (which is long enough to reach convergence).

\begin{figure}[ht]
\centering
    \includegraphics[width=0.85\textwidth]{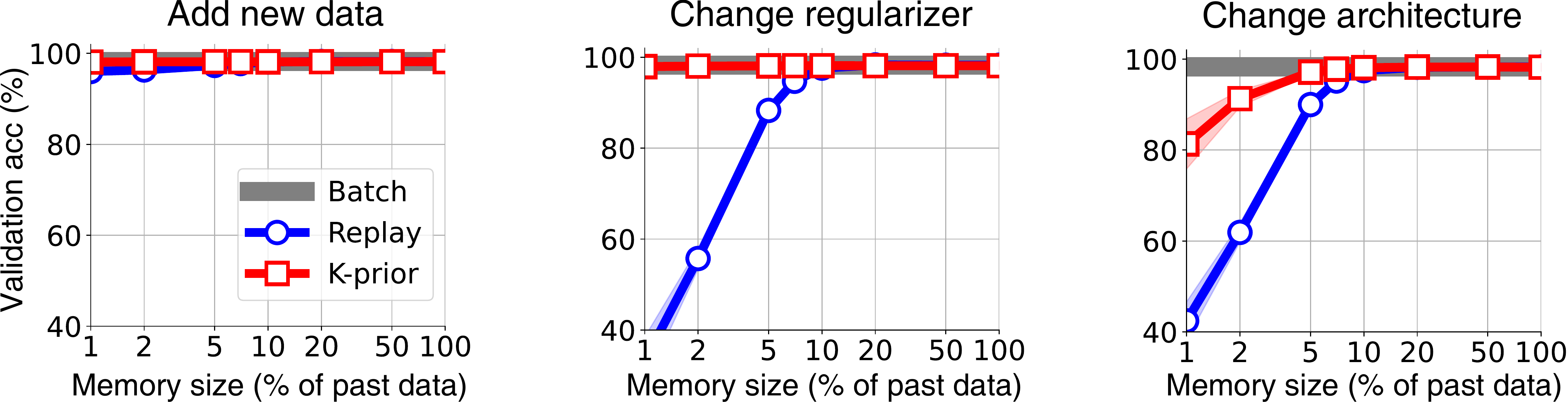}
    \caption{K-priors work well on MNIST (with an MLP), similar to other results on the USPS and UCI Adult datasets. For details on the experiments, see \cref{app:expts_adaptation}.}
    \label{fig:mnist_adaptation_tasks}
\end{figure}

\textbf{Neural Networks on the `MNIST' dataset.}
We show results on 10-way classification with MNIST in \cref{fig:mnist_adaptation_tasks}, which has 60,000 training images across 10 classes (handwritten digits), with each image of size $28\times28$. We use a two hidden-layer MLP with 100 units per layer, and report means and standard deviations across 3 runs. 
For the `Add Data' task, we start with a random 90\% of the dataset and add 10\%. 
For the `Change Regularizer' task, we change $\delta=1$ to $5$ (we use $\delta=1$ for all other tasks). 
For the `Change Architecture' task, we compress to a single hidden layer with 100 hidden units.
We optimize using Adam with a learning rate of $0.001$ for $250$ epochs, using a minibatch size of $512$. 

\textbf{Neural Networks on the `CIFAR-10' dataset.}
We provide results for CIFAR-10 using 10-way classification. CIFAR-10 has 60,000 images (50,000 for training), and each image has 3 channels, each of size $32\times32$.
We report mean and standard deviations over 3 runs.
We use the CifarNet architecture from \citet{zenke2017continual}.
We optimize using Adam with a learning rate of $0.001$ for 100 epochs, using a batch size of $128$. 

\begin{figure}[ht]
\centering
    \includegraphics[width=0.6\textwidth]{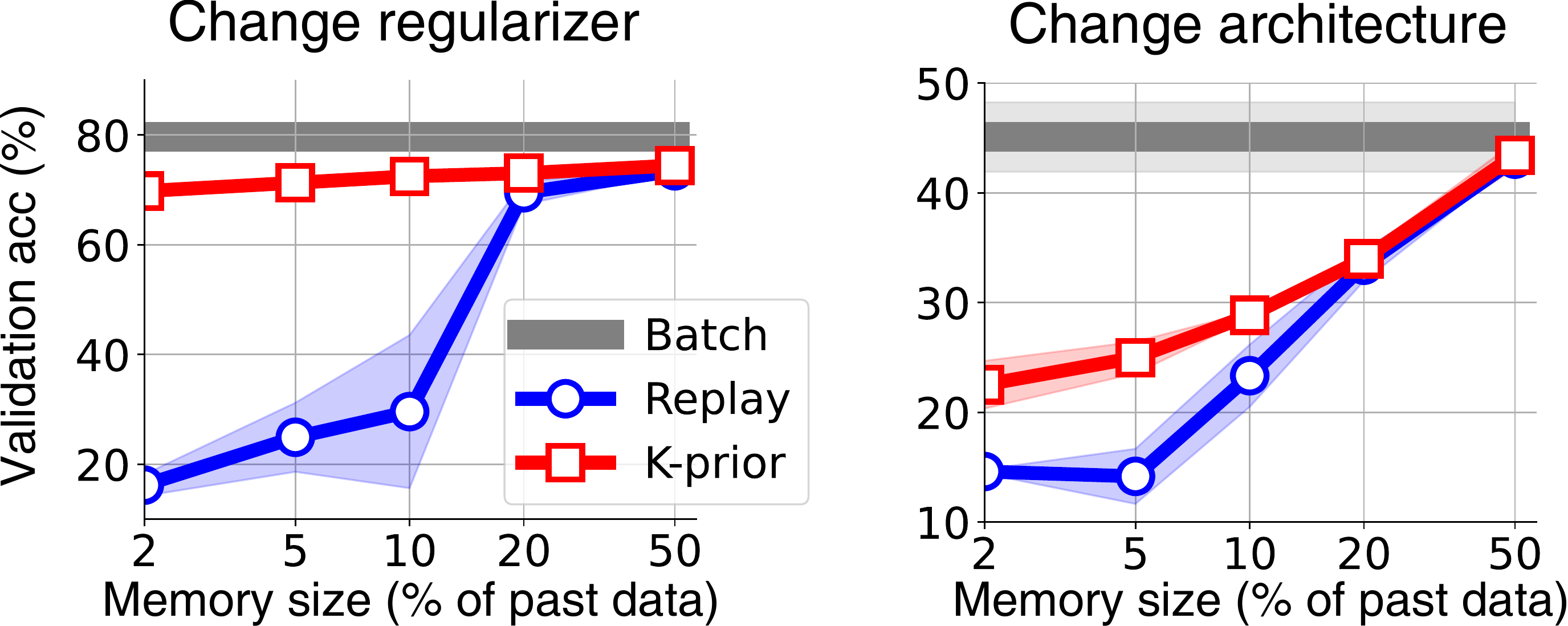}
    \caption{Results for two adaptation tasks on CIFAR-10 with CNNs. See also \cref{fig:fig3}(c) for results on the `Add Data' task. K-priors perform well, especially on the `Change Regularizer' task. The `Change Architecture' task is more difficult, but we note that we do not use a temperature. Having a temperature greater than 1 is known to help in similar settings, such as knowledge distillation \citep{hinton2015distilling}.}
    \label{fig:cifar_adaptation_tasks}
\end{figure}

In \cref{fig:cifar_adaptation_tasks} we also provide results on the `Change Regularizer' task, where we change $\delta=1$ to $0.5$ (we use $\delta=1$ for all the other tasks). 
We also provide results on the `Change Architecture' task, where we change from the CifarNet architecture to a LeNet5-style architecture. This smaller architecture has two convolution layers followed by two fully-connected layers: the first convolution layer has 6 output channels and kernel size 5, followed by the ReLU activation, followed by a Max Pool layer with kernel size 2 (and stride 2), followed by the second convolution layer with 16 output channels and kernel size 5, followed by the ReLU activation, followed by another Max Pool layer with kernel size 2 (and stride 2), followed by a fully-connected layer with 120 hidden units, followed by the last fully-connected layer with 84 hidden units. We also use ReLU activation functions in the fully-connected layers.

For the knowledge distillation task, we used K-priors with a temperature, similar to the temperature commonly used in knowledge distillation \citep{hinton2015distilling}. 
We note that there is some disagreement in the literature regarding how the temperature should be applied, with some works using a temperature only on the teacher's logits (such as in \cref{eq:KD}) \citep{lopez2015unifying}, and other works having a temperature on both the teacher and student's logits \citep{hinton2015distilling}.
In our experiments, we use a temperature $T$ on both the student and teacher logits, as written in the final term of \cref{eq:KD_expt}. 
We also multiply the final term by $T^2$ so that the gradient has the same magnitude as the other data term (as is common in knowledge distillation).
\begin{equation}
\begin{split}
   \loss_{\text{KD,expt}}(\vparam) = \lambda \sum_{i\in\data} \loss\rnd{ y_i, h(\modelout^i) }+ \delta\|\vparam\|^2 + (1-\lambda) \, T^2 \sum_{i\in\data} \ell\rnd{ h(\basemodel^i/T),\, h(\modelout^i/T) }.
   \label{eq:KD_expt}
\end{split}
\end{equation}
We used $\lambda=0.5$ in the experiment. We performed a hyperparameter sweep for the temperature (across $T=[1,5,10,20]$), and used $T=5$. For K-priors in this experiment, we optimize for $10$ epochs instead of $100$ epochs, and use $\tau=1$.

In \cref{fig:fig3}(c) we also showed initial results using a temperature on the `Add Data' task on CIFAR-10. We used the same temperature from the knowledge distillation experiment ($T=5$ and $\lambda=0.5$), but did not perform an additional hyperparameter sweep. 
We find that using a temperature improved results for CNNs, and we expect increased improvements if we perform further hyperparameter tuning. Note that many papers that use knowledge distillation perform more extensive hyperparameter sweeps than we have here.

\subsection{Weight-priors vs K-priors}

\begin{figure}[ht]
\centering
    \includegraphics[width=\textwidth]{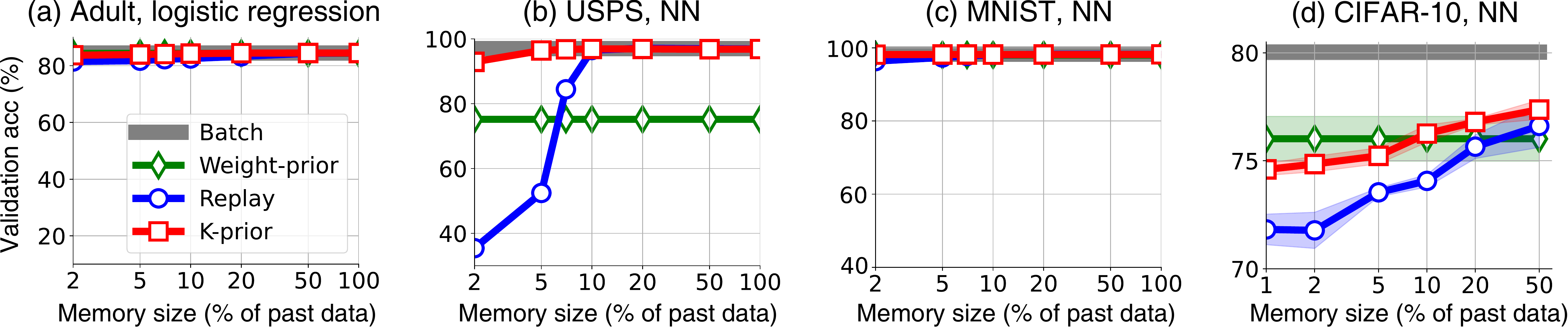}
    \caption{
    Results on the `Add Data' task, with a comparison to weight-priors. (a), (c), (d) For homogeneous data splits, weight-priors can perform relatively well. (b) For inhomogeneous data splits, weight-priors perform worse (see also \cref{fig:fig3}(b)).}
    \label{fig:add_data_weight-priors}
\end{figure}

In \cref{fig:add_data_weight-priors} we provide results comparing with weight-priors for all the `Add Data' tasks. We see that for homogeneous data splits (such as UCI Adult, MNIST and CIFAR), weight-priors perform relatively well. For inhomogeneous data splits (USPS with logistic regression and USPS with neural networks), weight-priors perform worse.

\subsection{K-priors ablation with weight-term}

\begin{figure}[ht]
\centering
    \includegraphics[width=0.75\textwidth]{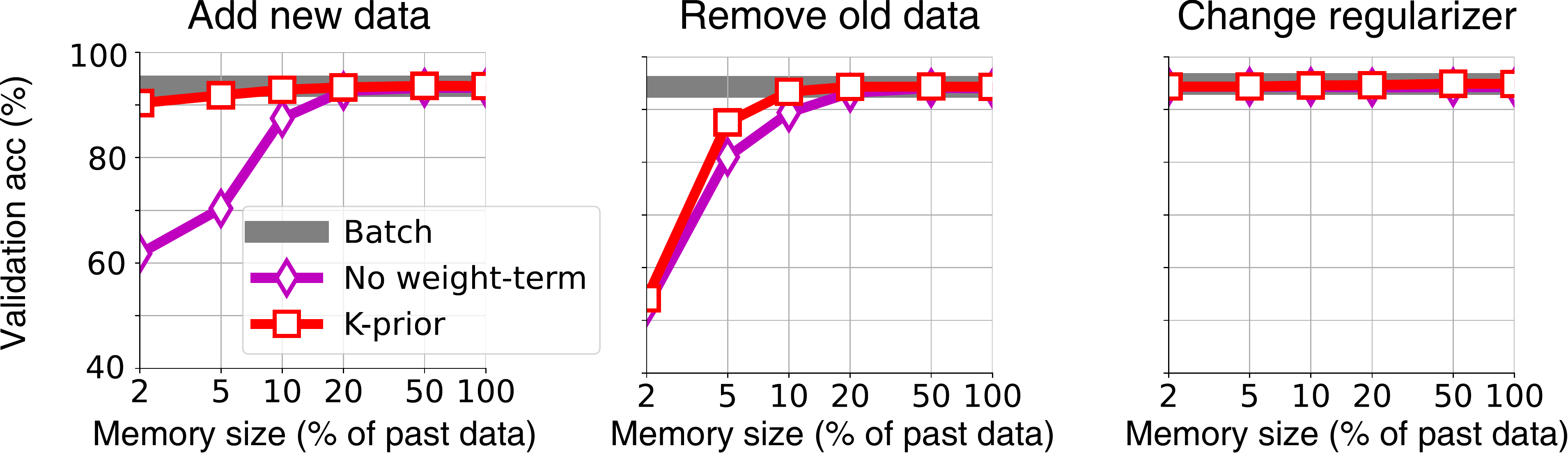}
    \caption{Comparing K-priors with a version of K-priors without the weight-term on USPS logistic regression. We see that the weight-term is important, especially on the `Add Data' task.}
    \label{fig:usps_no_weight_term_k_priors}
\end{figure}

In this section we perform an ablation study on the importance of the weight-term $\half \delta \| \vparam - \vparam_*\|^2$ in \cref{eq:kprior_glm}. 
In \cref{fig:usps_no_weight_term_k_priors} we show results on logistic regression on USPS where we do not have $\vparam_*$ in this term (the update equation is the same as \cref{eq:kprior_glm} except the weight-term is $\half \delta \| \vparam \|^2$ instead of $\half \delta \| \vparam - \vparam_*\|^2$).
We see that the weight-term is important: including the weight-term always improves performance.

\subsection{K-priors with random initialization}
\label{app:expt_rand_init}

In all experiments so far, when we train on a new task, we initialize the parameters at the previous parameters $\vparam_*$. Note that this is not possible in the ``Change architecture'' task, where weights were initialized randomly. Our results are independent of initialization strategy: we get the same results whether we use random initialization or initializing at previous values. The only difference is that random initialization can sometimes take longer until convergence (for all methods: Batch, Replay and K-priors). 

\begin{figure}[ht]
\centering
    \includegraphics[width=0.85\textwidth]{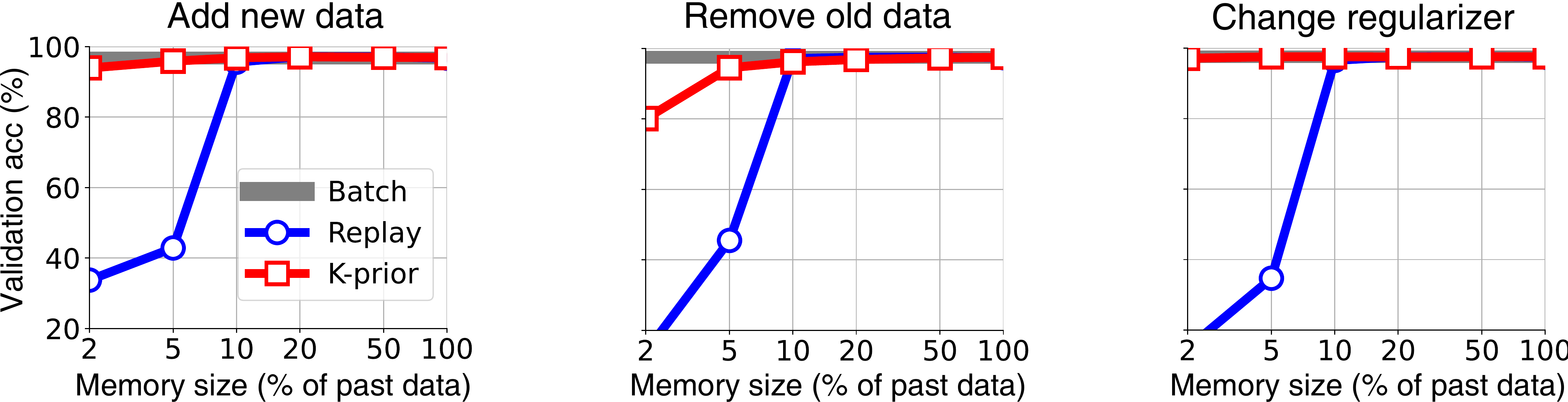}
    \caption{K-priors obtain the same results when randomly initializing the weights for the `Add new data', `Remove old data' and `Change regularizer' tasks on USPS odd vs even with neural networks. 
    Previous results, including \cref{fig:fig1} (right), initialized parameters at previously learnt values. 
    The `Change architecture' task originally used random initialization and so is not repeated here.}
    \label{fig:usps_random_init}
\end{figure}

For GLMs, where we always train until convergence and there is a single optimum, it is clear that the exact same solution will always be reached. We now also provide the result for `USPS odd vs even', with random initialization in \cref{fig:usps_random_init}, for the 3 tasks where we had earlier initialized at previous values (compare with \cref{fig:fig1} (right)). We use exactly the same hyperparameters and settings as in \cref{fig:fig1} (right), aside from initialization method.

\subsection{K-priors converge cheaply}
\label{app:expt_num_backprops}

In this section, we show that K-priors with limited memory converge to the final solution cheaply, converging in far fewer passes through data than the batch solution. This is because we use a limited memory, and only touch the more important datapoints.

\cref{tab:num_backprops} shows the “number of backprops” until reaching specific accuracies (90\% and 97\%) on USPS with a neural network (using the same settings as in \cref{fig:fig1} (right)). This is one way of measuring the ``time taken'', as backprops through the model are the time-limiting step. For K-priors and Replay, we use 10\% of past memory. All methods use random initializations when starting training on a new task.

We see that K-priors with 10\% of past data stored are quicker to converge than Batch, even though both eventually converge to the same accuracy (as seen in \cref{fig:fig1} (right)). For example, to reach 97\% accuracy for the Change Regularizer task, K-priors only need 54,000 backward passes, while Batch requires 2,700,000 backward passes. 
We also see that Replay is usually very slow to converge. This is because it does not use the same information as K-priors (as Replay uses hard labels), and therefore requires significantly more passes through data to achieve the same accuracy. 
In addition, Replay with 10\% of past data cannot achieve high accuracies (such as 97\% accuracy), as seen in \cref{fig:fig1} (right).

\begin{table*}[h]
    \centering
    \caption{Number of backpropagations required to achieve a specified accuracy on USPS with a neural network (1000s of backprops). K-priors with 10\% past memory require much fewer backprops to achieve the same accuracy as Batch, while Replay with 10\% memory cannot achieve high accuracies.
    }
    \begin{tabular}{llllll}
        \toprule 
        \textbf{Accuracy} & \textbf{Method} & \textbf{Add}      & \textbf{Remove}   & \textbf{Change}       & \textbf{Change}\\
        \textbf{achieved} &                 & \textbf{new data} & \textbf{old data} & \textbf{regularizer}  & \textbf{model class} \\
        \hline
            90\%    & Batch                 & 87  & 94  & 94  & 86  \\
            90\%    & Replay (10\% memory)  & 348 & 108 & 236 & 75  \\
            90\%    & \textbf{K-prior (10\% memory)} & \textbf{73}  & \textbf{53}  & \textbf{13}  & \textbf{22}  \\
            \hline
            97\%    & Batch                 & 1,900 & 1,800 & 2,700 & 3,124 \\
            97\%    & Replay (10\% memory)  & --    & 340   & --    & --  \\
            97\%    & \textbf{K-prior (10\% memory)} & \textbf{330}   & \textbf{120}   & \textbf{54}    & \textbf{68}  \\
        \bottomrule
    \end{tabular}
    \label{tab:num_backprops}
\end{table*}

\subsection{Further details on \cref{fig:fig1} (middle), moons dataset.}

To create this dataset, we took 500 samples from the moons dataset, and split them into 5 splits of 100 datapoints each, with each split having 50 datapoints from each task. Additionally, the splits were ordered according to the x-axis, meaning the 1st split were the left-most points, and the 5th split had the right-most points. In the provided visualisations, we show transfer from `past data' consisting of the first 3 splits (so, 300 datapoints) and the `new data' consisting of the 4th split (a new 100 datapoints). We store 3\% of past data as past memory in K-priors, chosen as the points with the highest $h'(\basemodel^i)$.

\section{Changes in the camera-ready version compared to the submitted version}

This section lists the major changes we made for the camera-ready version of the paper, incorporating reviewer feedback.

\begin{itemize}
    \item Added a paragraph on the optimal K-prior after \cref{eq:1order}, as well as a detailed explanation in \cref{app:kprior_svd}.
    \item Updated \cref{fig:fig3}(d), following a more extensive sweep of hyperparameters.
    \item Added \cref{app:expt_rand_init}, showing K-priors with random initialization give the same results as K-priors that are initialized at the previous model parameters.
    \item Added \cref{app:expt_num_backprops}, showing that K-priors with limited memory converge to the final solution cheaply, requiring fewer passes through the data than the batch solution.
\end{itemize}

\end{document}